\documentclass[11pt]{article}

\usepackage[final]{acl}

\usepackage{times}
\usepackage{latexsym}

\usepackage[normalem]{ulem}
\usepackage{soul}

\usepackage{amsmath}
\usepackage{multirow}

\usepackage[T1]{fontenc}

\usepackage[utf8]{inputenc}

\usepackage{microtype}

\usepackage{inconsolata}

\usepackage{graphicx}
\usepackage{subcaption}

\usepackage{booktabs}
\usepackage{multirow}
\usepackage{pdflscape} 
\usepackage{makecell}  
\usepackage{diagbox} 
\usepackage{adjustbox}
\usepackage{colortbl}

\usepackage{soul}

\definecolor{lightyellow}{rgb}{1.0, 1.0, 0.6}

\usepackage[utf8]{inputenc}
\usepackage[english]{babel}
\usepackage{ucs}

\definecolor{lightyellow}{rgb}{1.0, 1.0, 0.6}

\title{Multilinguality Does not Make Sense: Investigating Factors Behind Zero-Shot Cross-Lingual Transfer in Sense-Aware Tasks
}

\author{Roksana Goworek$^{\mathbf{1}}$ Haim Dubossarsky$^{\mathbf{1}, }$$^{\mathbf{2},}$$^{\mathbf{3}}$  \\
$^{\mathbf{1}}$ Queen Mary University of London \\
$^{\mathbf{2}}$ The Alan Turing Institute\\
$^{\mathbf{3}}$ University of Cambridge\\
\texttt{\{r.goworek, h.dubossarsky\}@qmul.ac.uk}  }

\begin{document}
\maketitle
\begin{abstract}


Cross-lingual transfer allows models to perform tasks in languages unseen during training and is often assumed to benefit from increased multilinguality. In this work, we challenge this assumption in the context of two underexplored, sense-aware tasks: polysemy disambiguation and lexical semantic change. Through a large-scale analysis across 28 languages, we show that multilingual training is neither necessary nor inherently beneficial for effective transfer. Instead, we find that confounding factors, such as fine-tuning data composition and evaluation artifacts, can better account for the perceived advantages of multilinguality. Our findings call for more rigorous evaluations in multilingual NLP, and more nuanced and sensible choice of models for transfer. We release fine-tuned models and benchmarks to support further research, with implications extending to low-resource and typologically diverse languages.

\end{abstract}

\section{Introduction}


Cross-lingual transfer enables multilingual pretrained models to leverage knowledge acquired in one language to perform tasks in another (e.g., \citealp{wu2019beto, ponti2018adversarial}). This ability underpins many of today’s advances in multilingual NLP and has been evaluated across a wide range of tasks - from syntactic parsing and POS tagging to more complex, semantics-driven tasks like question answering, language inference, and paraphrasing (see \citet{philippy-etal-2023-towards} for a review).

Polysemy disambiguation, despite being foundational to linguistic meaning and a long-standing challenge in NLP \citep{navigli2009word, bevilacqua2021recent}, has received relatively little attention in transfer learning research. Polysemy poses unique challenges to transfer due to its inherently language-specific nature \citep{rzymski2020database}. For example, while the English word mole denotes ‘a small burrowing mammal’ and ‘a skin blemish,’ its Hindi counterpart \includegraphics[height=0.9em]{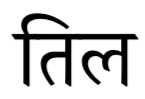} (til) refers to the latter sense but also means ‘sesame seed’. Such non-aligned senses across languages complicate direct transfer, making it an ideal testbed for evaluating the true limits of cross-lingual generalization.

A common assumption is that multilinguality itself, set by a model's exposure to multiple languages, is key for cross-lingual transfer. While it is evident that models must support both source and target languages, the extent to which broader multilinguality facilitates transfer remains unclear and understudied. Does access to more languages intrinsically improve transfer, or do other factors, such as language similarity or training setup, drive the observed gains?

In this work, we investigate the role of multilinguality in cross-lingual transfer for two underexplored, \textbf{sense-aware tasks}: polysemy disambiguation and lexical semantic change detection. We examine whether multilingual fine-tuning is truly essential for successful transfer, or whether previously reported benefits stem from confounding factors. Our results show that multilinguality is neither necessary nor intrinsically beneficial in these settings, challenging prevailing assumptions about the mechanisms underlying multilingual transfer.

While our findings may seem narrowly focused, they illuminate broader issues central to NLP, with implications for low-resource and domain-specific settings for which relying on broader multilingual models and corpora is not an option, as well as the wider research community. We release the best-performing models to support further analysis.\footnote{Best-performing \textsc{mono} and \textsc{multi} models are available in the collection
~\href{https://huggingface.co/collections/Roksana/multilinguality-does-not-make-sense-68cd7ccc903a72b3af2f978a}{Multilinguality Does not Make Sense}.}

\section{Related Work}
\label{sec:related_word}

Cross-lingual transfer studies have expanded from tasks like syntactic parsing, POS tagging, and semantic classification \cite{wu2019beto, choenni-etal-2023-cross, pires2019multilingual, de-vries-etal-2022-make} to more complex ones such as NER, NLI \cite{dolicki2021analysing, srinivasan2021predicting}, and more recently QA, paraphrasing, and sentiment analysis \cite{lauscher2020zero, ahuja-etal-2023-mega, choenni-etal-2023-languages, wang-etal-2023-self-augmentation}. Most studies rely on pretrained multilingual models like XLM-R and mBERT.

Research also focused on factors affecting transfer, including linguistic similarity \cite{lauscher2020zero}, pretraining corpus size and diversity \cite{srinivasan2021predicting, ahuja-etal-2023-mega}, lexical overlap \cite{patil-etal-2022-overlap, de-vries-etal-2022-make}, and model architecture \cite{K2020Cross-Lingual}. Language selection also varied, from high-resource \cite{choenni-etal-2023-languages} to extremely low-resource languages tackled with continuous pretraining \cite{ebrahimi-kann-2021-adapt, imanigooghari-etal-2023-glot500}.

Despite extensive work, the role of multilinguality itself is rarely tested directly and is often assumed to be beneficial. A notable exception is \citet{shaham2024multilingual}, who compare PaLM 2 fine-tuned on monolingual versus multilingual data for instruction tuning, and find that modest multilingual exposure aids transfer, while too much can degrade it. However, their analysis is limited to a single model and task, leaving open questions about pretraining and generalizability. \citet{choenni-etal-2023-languages} show that source language examples can influence target predictions, but without explicitly controlling for multilinguality, offering only indirect evidence. \citet{chang-etal-2024-multilinguality} directly manipulated multilinguality by pretraining over 10,000 models across 250 languages, but only evaluated using perplexity with no standard tasks, making comparisons difficult and replication impractical.

Notably underrepresented in transfer research are sense-aware tasks, such as polysemy disambiguation and lexical semantic change (LSC), despite the central role polysemy has in NLP. Unlike sentiment analysis or other tasks amenable to meaning-preserving translations, polysemy exhibits substantial language-specific variation \cite{rzymski2020database}, making it particularly suitable for rigorous evaluation of zero-shot multilingual transfer ability. 

Exceptions for polysemy are few. \citet{raganato2020xl} reported transfer to 12 languages, but used only English as a source language, while \citet{dubossarsky2024strengthening} found near-zero transfer to Hindi, raising questions about the feasibility of zero-shot transfer in low-resource settings. In contrast, \citet{goworek-etal-2025-senwich} recently showed notable zero-shot transfer from English to 10 low-resource languages, highlighting the need for further investigation.

With regard to LSC, \citet{arefyev-2021-etal} showed that training on polysemy disambiguation generalizes to semantic change detection, linking the two tasks. \citet{cassotti2023xl} followed up and fine-tuned models on multilingual polysemy datasets, achieving state-of-the-art results \cite{schlechtweg2020semeval} and strong transfer to unseen languages \cite{periti2024systematic}.

None of these works studied multilinguality itself. A notable exception is \citet{berend-2022-combating}, who explicitly examined the role of multilinguality in word sense disambiguation transfer. However, their study investigated the role of multilinguality only at the pretraining stage, rather than during fine-tuning. This is a less practical perspective given the ubiquitous use of multilingual models in NLP today, where many languages lack high quality monolingual models. 



Overall, while multilingual transfer has been widely studied, the direct contribution of multilinguality in fine-tuning remains underexplored, especially in lexically focused tasks like polysemy and LSC. This work addresses that gap by systematically testing the role of multilinguality in sense-aware transfer. By manipulating multilingual conditions, we clarify when and how multilinguality supports cross-lingual generalization, helping to resolve prior conflicting findings and inform future research on transfer learning.

\section{Methods}

In this study, we set to isolate
multilinguality from confounds, enabling a clear
assessment of its independent impact on zero-shot
transfer in polysemy disambiguation and LSC tasks.
We conduct a large-scale evaluation of zero-shot and full-shot cross-lingual transfer performance across 28 languages, focusing on the unique contribution of multilingual training. To achieve this, we implement an experimental framework that systematically controls for potential confounds, ensuring that observed effects are attributed to multilinguality rather than artifacts of training data size or pretraining exposure.

\subsection{Tasks}
\label{subsect:taskWIC}
The Word in Context (WiC) formulation of polysemy disambiguation is used. By pairing two sentences with the same polysemous word \citet{pilehvar2018wic} transformed this task into a binary classification problem, where a target word appears either in the \textit{same sense} or in a \textit{different sense} as per the example below:
\begin{quote}
    \vspace{-4pt} 
    The couple went for a \textbf{date} last night. \newline
    He marked this \textbf{date} on my calendar.
    \vspace{-4pt}
\end{quote}


\noindent This reformulation removed the need for a sense-label per word, which is language-specific, making the task more suitable for cross-lingual transfer.



For LSC, we compare the models' representations of words occurring in natural sentences across two corpora from different time periods. The underlying assumption, which is the basis of all distributional semantics, is that changes in words' meaning are reflected in measurable changes to their representations over time \cite{periti2024lexical}.

\subsection{Data}
\label{subsect:wicDatasets}
MCL, XL and Hindi datasets, which together span 18 languages, were used for training and evaluation, while AM$^2$iCO and LSCD (LSC Detection), spanning 14 and 7 languages, respectively, were used only for evaluation.\footnote{AM$^2$iCO uses English as a pivot language therefore is biased toward it; LSCD is not in the WiC format.} As some languages have only development and test data, we followed \citet{cassotti2023xl} who sampled 75\% of the development data of each language to enable training on these languages (keeping the remaining 25\% for setting hyperparameters), using their exact train-dev-test splits. All WiC datasets are class-balanced, setting the chance baseline at 50\%. German, French and English are overrepresented in the dataset relative to other languages (see Figure~\ref{fig:pie_chart}).



\textbf{MCL} by \citet{martelli2021semeval} spans English, Arabic, French, Russian and Chinese,
constructed by annotating sentences from native corpora: BabelNet \cite{navigli-ponzetto-2010-babelnet}, the United Nations Parallel Corpus \cite{ziemski-etal-2016-united} and Wikipedia, with inter-annotator agreement of 0.95 and 0.9 on English and Russian, respectively.

\textbf{XL} by \citet{raganato2020xl} used WordNet of Bulgarian, Chinese, Croatian, Danish, Dutch, English, Estonian, Japanese, Korean and Farsi, filtering out fine-grained senses. French, German and Italian used Wiktionary. The reported mean human accuracy was 80\%, and varied between 74\% for German, 87\% for Danish and 97\% for Farsi.

\textbf{Hindi WiC} by \citet{dubossarsky2024strengthening} consists of 12,000 sentence pairs, constructed from a sense-annotated Hindi Corpus \cite{singh2016sense} of 60 polysemous nouns. 

\textbf{AM$^2$iCO} by \citet{liu2021am2ico} has English paired with 14 target languages. Compiled from Wikipedia Dumps of each language by selecting words with at least two different Wikipedia pages that show ambiguity in both the target language and English.
Overall human accuracy was 90.6\%, with an inter-annotator agreement of 88.4\%.

\textbf{Lexical Semantic Change Detection (LSCD)} 
\label{sec:LSCD_method}
covers seven languages from different sources: English, German, Latin, and Swedish from \citet{schlechtweg2020semeval}, Spanish from \citet{zamora2022lscdiscovery}, Chinese from \citet{chen2023chiwug}, and Norwegian from \citet{kutuzov2022nordiachange}.\footnote{Norwegian$_1$, Norwegian$_2$, are two corpus pairs, comparing the words from four time periods.} For each language and target word, an equal number of sentences were sampled from corpus 1 (historical) and corpus 2 (modern).




\label{section: basemodels}
\subsection{Multilingual Base Models}
\label{subsect:models}

To ensure robustness, we use five multilingual models that differ in their language coverage, and in the pretraining proportions of these languages. 

\textbf{XLM-R-large} \cite{conneau2019unsupervised} and \textbf{mBERT}  \cite{devlin2018bert} are pretrained on 100 and 104 languages, respectively, covering all languages in our study, which enables us to train and evaluate on all languages in zero- and full-shot transfer. Both models are commonly used in multilingual research.
XLM-R was extensively used in the context of WiC and LSC \citep{raganato2020xl, cassotti2023xl, dubossarsky2024strengthening}, providing a strong baseline for comparison. 

\textbf{BLOOM} (BigScience Large Open-science Open-access Multilingual Language Model) by \citet{le2023bloom} is pretrained on 46 languages, with a focus on low-resource languages, 10 of which overlap with those used in our study. 

\textbf{LLaMA3-8B-Instruct} \cite{grattafiori2024llama, llama3modelcard} is a decoder-only instruction-tuned language model with 8 billion parameters, significantly larger than the other models. Pretraining data is not public, and is assumed to include major NLP datasets \cite{sainz-etal-2023-nlp}, like WiC.

\textbf{MuRIL} (Multilingual Representations for Indian Languages) by \citet{khanuja2021muril} is pretrained on 16 Indian languages + English, which, along with Hindi, are the only languages overlapping with our study. It was used to test an edge case of zero-shot transfer (see \S\ref{subsec:muril results}).

\begin{figure*}[h]
    \centering
    \includegraphics[width=\linewidth]{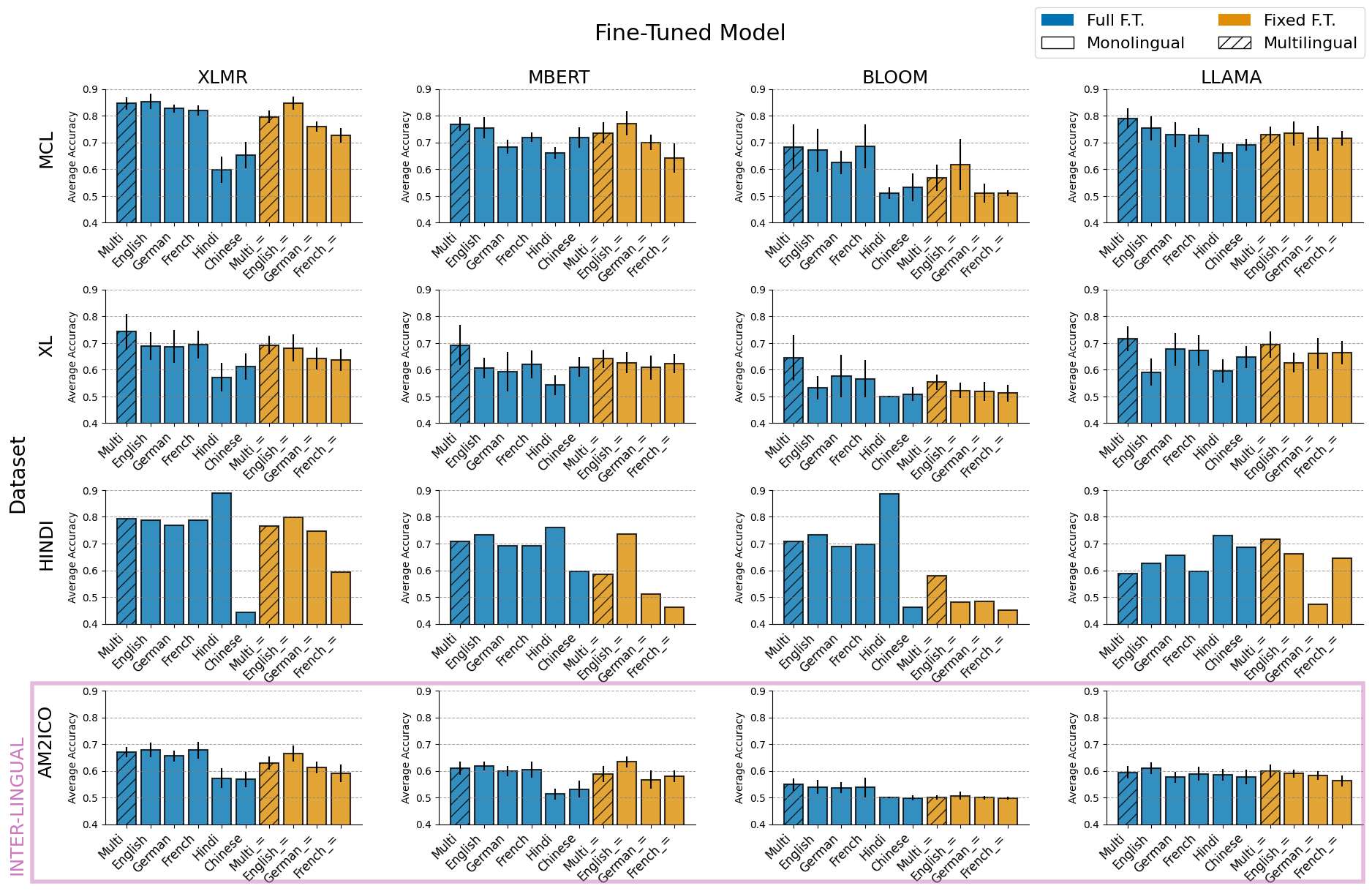}
    \caption{
    Mean accuracies and \textsc{sd} (bars) for multilingual and monolingual models on WiC datasets using different pretrained models. Colors indicate whether fine-tuning was done on all data or on its sampled portion (Hindi and Chinese appear only in the former due to their smaller data which did not allow subsampling). Hindi, as a single language dataset, does not have \textsc{sd}. For detailed results see Appendix~\ref{app:all_results}. }
    \label{fig:mlcl_vs_mono_ft}
\end{figure*}

\subsection{Model Training and Testing}

We fine-tune models on different combinations of the data, creating three conditions: (1) \textsc{\textbf{mono}}lingual models, trained on a single language. Only the 5 largest languages were considered; (2) \textsc{\textbf{multi}}lingual models, trained on all languages; and (3) \textsc{\textbf{multi$_{\otimes}$}}lingual models, trained on all but one held-out language. 
We also compare these models while controlling for the total amount of fine-tuning in the \textsc{fixed f.t.} condition, denoted by \textsc{\textbf{model}$_=$}. By systematically comparing performance across conditions while controlling for fine-tuning size and pretraining exposure, we isolate the effect of multilinguality on cross-lingual transfer in WiC and LSC tasks.

\paragraph{Fixed Fine-Tuning}
\label{subsect:Fix_FT}
To control for the different training sizes across languages, fixed-size versions of datasets were created for the \textsc{multi}, \textsc{english}, \textsc{french} and \textsc{german} datasets by randomly subsampling 8,750 examples from their training sets.\footnote{Other languages had too little data to subsample.}

\paragraph{Finetuning Details} All encoder models were fine-tuned in the same way, selectively focusing on different subsets of the WiC datasets. 
Following \citet{cassotti2023xl} and after a failed pilot study with cross-encoders, a Siamese bi-encoder was used to generate two distinct vector representations (embeddings) for the two usages of the target word in the two sentences.
The model outputs the cosine distance between the output embeddings of the two inputs, and, in order to adapt it to the binary nature of WiC, a threshold is applied to decide if the words are classified as having the same sense. The model is trained to update its parameters (i.e., embeddings) to maximize this distance when the target appears in different meanings (label 0) and minimize it when the meanings are the same in the two sentences (label 1), by minimizing contrastive loss.
During training, as well as inference, special tokens, \texttt{<t>} and \texttt{</t>}, are placed around the target word in each sentence to signal what word the model should focus on. 

\interfootnotelinepenalty=10000

To preserve comparability and avoid cherry-picking, we fixed a single hyperparameter configuration and did not perform per-model/per-language tuning or multi-seed sweeps, which may understate best-case performance and limit our characterization of variance across runs. For more discussion of this decision, see Appendix~\ref{app:hyperparameter_justification}.\footnote{
Training hyperparameters and model sizes are in Table~\ref{tab:hyperparameters}.} 

To verify that performance is significantly improved by fine-tuning rather than being the result of inherent sense-distinction ability of contextualized embeddings, we conduct no-fine-tuning experiments with frozen pretrained models. The results corroborate this and can be found in Appendix~\ref{app:paramfree}.
For details of LLaMA’s fine-tuning (same data, different procedure), see Appendix~\ref{sec:llama_details}.

\paragraph{Testing} To evaluate models on WiC, a threshold for each model is determined by maximizing accuracy on the validation set of the training language. This threshold collapses the cosine distance between the output embeddings of the two inputs into a binary label. For LSC, we follow \citet{cassotti2023xl} and output a fixed-size vector for the target word in each sentence. Taking all vectors for each word across the two time-bins, a change score is computed using APD and PRT measures \cite{kutuzov2020uio}, and evaluated against gold-label change scores using Spearman correlation, as standard in LSC \cite{periti2024systematic}.


    


\section{Experiments and Evaluation}



\subsection{Multilinguality Effects and Confounds}
\label{subsect:mulilingualConfounds}
To rigorously assess the role of multilinguality in zero-shot transfer, we conduct three types of controlled comparisons, ensuring that fine-tuning size is held constant and distinguishing between full-shot and true zero-shot transfer scenarios. We provide a comprehensive evaluation of four multilingual base-models (\S\ref{subsect:models}) on 28 languages, first examining whether multilingual training offers an advantage and then systematically accounting for key confounding factors. We summarize the average accuracies of the models in Figure~\ref{fig:mlcl_vs_mono_ft} and present the true zero-shot transfer comparison in Table~\ref{tab:zero-shot}. Due to space constraints, except for Figure~\ref{fig:mlcl_vs_mono_ft}, we show and discuss XLM-R results in the main text, only briefly discussing other models while providing their full results in Appendix~\ref{app:all_results}.


%

\paragraph{Testing for Multilingual Advantage}
We contrast between \textsc{mono} models, which are fine-tuned on a single language, and the \textsc{multi} model, trained on all available data in the MCL and XL WiC datasets. If multilinguality is indeed critical, then \textsc{multi} should consistently outperform \textsc{mono} models, which by definition do not have access to information outside of their own language beyond their pretraining stage. Our results show a mixed pattern, with many \textsc{mono} models on-par or even better than \textsc{multi} on the MCL, AM$^2$iCO and HINDI datasets, and a multilinguality advantage on the XL dataset (blue bars of the \textsc{Full F.T} condition in Figure~\ref{fig:mlcl_vs_mono_ft}). 
LLaMA is the exception to the rule, though its results are 5\%-10\% points lower than XLM-R. We attribute the low transfer of Hindi and Chinese \textsc{mono} models to their smaller dataset sizes, and subsequently tested whether dataset size is a potential confound in our analyses. 




\paragraph{Dataset Size Confound}

\textsc{multi} is fine-tuned on an order of magnitude more data than any other model (see Figure \ref{fig:pie_chart}), which gives it an unfair advantage over all monolingual models. 
We therefore fix the size of the training datasets across all models (\S\ref{subsect:Fix_FT}) to enable a fair comparison, and repeat the same analysis under these controlled conditions. The results in Figure~\ref{fig:mlcl_vs_mono_ft}, and especially the comparison between \textsc{Full F.T} and \textsc{Fixed F.T} (blue and orange bars, respectively) which are summarized in Table~\ref{tab:ft_drop_in_perf}, show that the performance of the \textsc{multi} model drops much more than \textsc{mono} English when training size is controlled, perhaps due to a larger relative drop in training size, making \textsc{mono} English the best model across most datasets. Similar drops are observed for mBERT, BLOOM, and LLaMA (Tables \ref{tab:mbert_ft_drop_in_perf}-\ref{tab:llama_ft_drop_in_perf}), which for the latter largely diminished the advantage it had in \textsc{Full F.T}.
Despite its drop in the \textsc{Fixed F.T} condition, LLaMA seems to benefit from multilingual fine-tuning more than other models. We attribute this to its nature as a generative model and pretraining uncertainty which may include data contamination unwanted for our experiments \cite{sainz-etal-2023-nlp}. Additionally, LLaMA relies on English prompts with explicit task instructions, which showed improved performance, particularly for \textsc{hindi} and \textsc{chinese}, where training data is more limited.
Overall, this contrast (blue and orange) undermines the notion that training on a multilingual dataset inherently improves transfer. Instead, the advantage originally observed for \textsc{multi} in the Full F.T may largely be due to more training data (122.4k examples for \textsc{multi} cf. 54.7k for German, 46.1k for French, and 15.1k for English; see Table \ref{tab:model_description}). This finding does not imply that training on more examples is not a good strategy to improve transfer, which it clearly is, only that attributing the performance gains to multilinguality is flawed.

\paragraph{Error Analysis}

If multilinguality equips models with novel sense-understanding, then not only should its performance improve, but also its errors should differ from \textsc{mono} models that lack exposure to such linguistic diversity. Instead, Figure~\ref{fig:incorrect_mono_alignment} shows  greater overlap of errors between \textsc{multi} and \textsc{mono} models, which even increases for \textsc{Fixed F.T}, supporting a notion that \textsc{multi} is simply another \textsc{mono} model (see formulation of alignment measures in Appendix~\ref{sec:error_analysis}). This provides converging evidence that supports previous findings, further undermining the importance of multilinguality.

\begin{table}[!ht]
    \centering
    \small
    \begin{tabular}{|l|l|l|l|l|}
    \hline
       \diagbox{Model}{Dataset} & MCL & XL & Hindi & Am$^2$iCO \\ \hline
        \textsc{multi} & -5.0 & -5.0 & -2.7 & -4.0 \\ \hline
        \textsc{english} & -0.6 & -0.6 & 1.2 & -1.5 \\ \hline
        \textsc{german} & -6.8 & -4.4 & -2.0 & -4.3 \\ \hline
        \textsc{french} & -9.5 & -5.5 & -19.6 & -8.7 \\ \hline
    \end{tabular}
    \caption{Percentage change in average accuracy of XLM-R per dataset (Fixed F.T. - Full F.T.)}
    \label{tab:ft_drop_in_perf}
\end{table}


    
    
    

    
    
    
    
\begin{table}[h]
\centering
\tiny
\resizebox{\columnwidth}{!}{
\begin{tabular}{|c|c|c|c|c|c|c|c|c|}
\hline
\raisebox{-3.5ex}{Language}
& \rotatebox{270}{German  }
& \multicolumn{2}{c|}{\rotatebox{270}{French}}
& \multicolumn{2}{c|}{\rotatebox{270}{English}}
& \rotatebox{270}{Hindi}
& \multicolumn{2}{c|}{\rotatebox{270}{Chinese}} \\ \hline

    Dataset & \rotatebox{0}{XL} 
    & \rotatebox{0}{MCL} 
    & \rotatebox{0}{XL} 
    & \rotatebox{0}{MCL} 
    & \rotatebox{0}{XL} 
    & \rotatebox{0}{Hindi} 
    & \rotatebox{0}{MCL} 
    & \rotatebox{0}{XL} \\ \hline \hline
\textsc{z.s. multi$_{\otimes=}$} & 67.6   & 79.0         & 65.3        & \textbf{82.9}     & \textbf{68.7}     & 76.6  & 78.3         & 72.2         \\ \hline \hline
\textsc{german$_=$}             & \cellcolor{gray} 74.1   & 77.0         & 63.8        & 78.3           & 60.6         & 74.7  & 75.0         & 67.5         \\ \hline
\textsc{french$_=$}             & 68.6   & \cellcolor{gray} 73.7   & \cellcolor{gray} 65.5        & 77.0           & 63.0         & 59.3  & 71.2         & 66.9         \\ \hline
\textsc{english$_=$}            & \textbf{69.3}  & \textbf{84.1}        & \textbf{65.6}       & \cellcolor{gray} 89.2  & \cellcolor{gray} 68.9  & \textbf{79.9} & \textbf{78.4}        & \textbf{76.2}        \\ \hline
\textsc{hindi$_=$}              & 62.1   & 53.8         & 59.7        & 66.0           & 56.8         & \cellcolor{gray} 88.9  & 60.7         & 66.4         \\ \hline
\textsc{chinese$_=$}            & 62.3   & 61.3         & 58.1        & 68.1           & 61.3         & 44.2  & \cellcolor{gray} 70.8  & \cellcolor{gray} 69.4  \\ \hline
\end{tabular}}
\caption{XLM-R (fixed F.T.) \textbf{zero-shot} results for \textsc{mono} and \textsc{multi} models.
For each of the 5 column languages, the \textsc{z.s.} \textsc{multi}$_{\otimes=}$ row reports the model's performance when it is trained multilingually with that language held out.
Rows \textit{lang}$_{=}$ report  monolingual models trained on \textit{lang}, evaluated on each column test language.
Grey cells indicate full-shot conditions (target language included in fine-tuning).}

\label{tab:zero-shot}
\end{table}

The \textsc{z.s. multi$_{\otimes=}$} row presents the results of five different multi$_{\otimes=}$ models (each trained on all available languages excluding the target language, specified by the column, respectively). The \{lang\}$_{=}$ rows show the results of the specified monolingual model on the test set specified by the column. Grey cells  mark full-shot conditions for monolingual models.



\paragraph{Full-shot Exposure Artifact}
We currently measure transfer by conflating zero- and full-shot, when the target language is present in training, together. Since \textsc{multi} is trained on all the languages it is later evaluated on (except Hindi), it is effectively evaluated under full-shot conditions. In contrast, \textsc{mono} models are only full-shot with respect to their training language, and zero-shot for all the rest. Thus, comparing transfer between \textsc{multi} and \textsc{mono} models is unfair, as full-shot learning is expected to be much better than zero-shot transfer.

To address this, we train five different \textsc{multi$_{\otimes}$} models, excluding one language at a time on which that specific model is later evaluated on, while still fixing the amount of fine-tuning data as before, and compare them to \textsc{mono} models. We focus on the 5 languages with the most training data. If multilinguality is truly of merit, then \textsc{multi} models, that were trained on 14 languages overall (minus the held-out language they are evaluated on), should outperform \textsc{mono} models. 

Table~\ref{tab:zero-shot} shows that \textsc{mono} English outperforms zero-shot \textsc{multi$_{\otimes}$} on all languages, and the only case where zero-shot \textsc{multi$_{\otimes}$} outperforms \textsc{mono} models is English, where the \textsc{english} model is full-shot, and thus not considered in this evaluation (grayed). This further disproves the assumed benefit of multilingual training, as prior advantage in performance that was originally associated with multilinguality stemmed, at least in part, from mixing full-shot with zero-shot transfer conditions, which only \textsc{multi} possessed. Similar results were obtained for mBERT, BLOOM and even LLaMA, although with \textsc{german} as the best zero-shot model (see Tables \ref{tab:zero-shot_mBERT}-\ref{tab:zero-shot_LLaMA}).
Interestingly, the strong zero-shot performance of \textsc{multi} models on English could be related to the prevalence of English in the base models' pretraining data.

\subsection{Underlying Factors of Successful Transfer}

\paragraph{Correlation with Model's Pretraining Size}
\label{sec:pretraining_size}

Our analyses reject multilinguality as a key driver of zero-shot transfer, showing its effects stem from training data size - a confound, albeit a beneficial one. Yet, even after controlling for this, transfer results varied considerably across target languages.

Before their fine-tuning, multilingual models undergo pretraining on large-scale datasets spanning multiple languages. The prevalence of a target language within a model's pretraining corpus, or "pretraining size", may influence the model's ability to represent, process and transfer to that language. 

To test this hypothesis, and understand what drives variation in transfer results, we computed Pearson correlations between languages' log-transformed pretraining sizes in models for which pretraining sizes were available (XLM-R, mBERT, BLOOM) and their corresponding accuracies on the four WiC datasets and on LSCD.\footnote{Pretraining sizes are provided in Appendix~\ref{tab:pretraining_sizes}.
For BLOOM, we removed languages not in its pretraining.}

\begin{figure}[h]
    \centering
    \includegraphics[width=\linewidth]{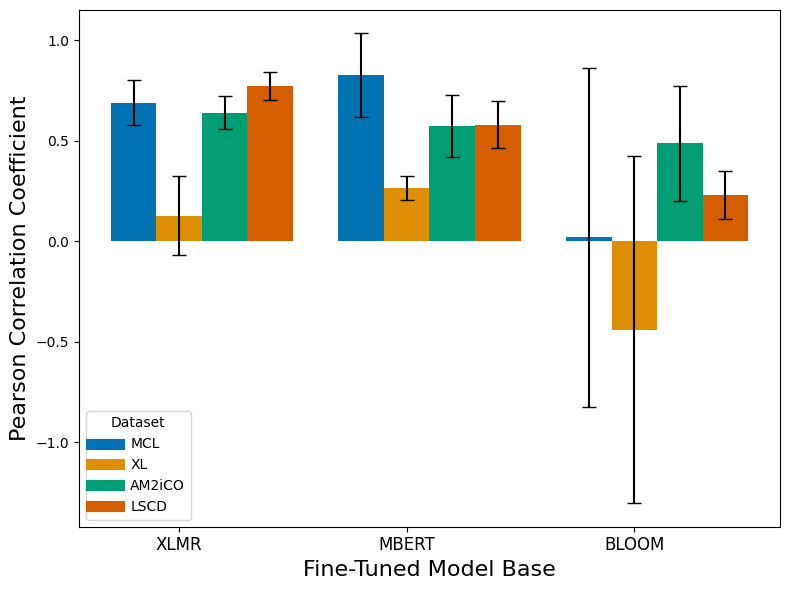}
    \caption{Mean correlations between languages pretraining sizes and zero-shot performance of \textsc{mono} models.
    }
    \label{fig:pt-corr}
\end{figure}

Figure \ref{fig:pt-corr} shows strong correlations between a language's pretraining size and its accuracy as a target language. 
We attribute BLOOM's poor correlations to its unusual pretraining language distribution that focuses on low-resource languages. The lack of correlation in XL remains unclear and requires further investigation (see \S\ref{sec:dicussion} for discussion). 

\paragraph{Linguistic Similarity} 
\label{sec:linguistic_similarity}
We tested the link between linguistic similarity and transferability using syntactic similarity scores \cite{littell-etal-2017-uriel}, which quantify similarity based on shared syntactic features, as is standard in NLP \cite{philippy-etal-2023-towards}. 
Pearson correlations between similarity scores and model accuracies were computed across target language pairs. Syntactic similarity showed some correlation with zero-shot performance, the effect was weaker and less consistent than that of pretraining size (see Appendix~\ref{fig:syntactic_similarity_corr}). Moreover, we found that syntactic similarity is highly correlated with pretraining size, further undermining its contribution.

\begin{figure}
    \centering
    \includegraphics[width=\linewidth]{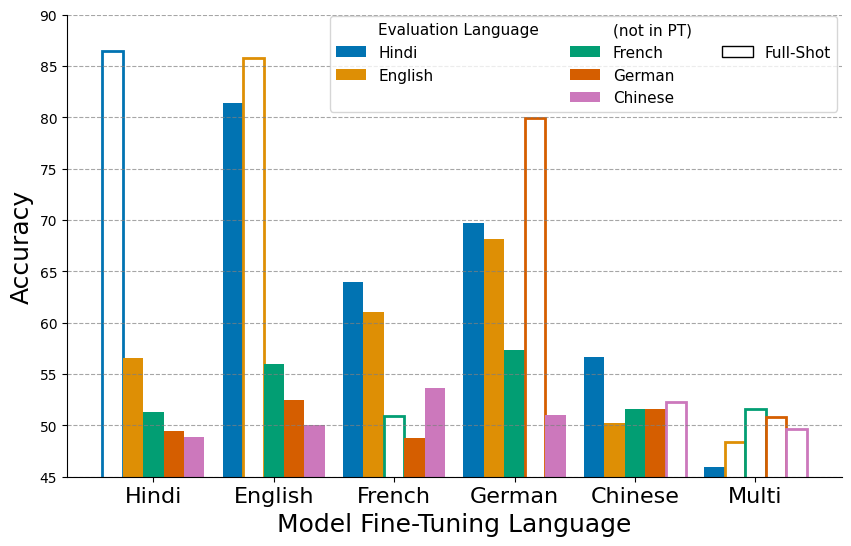}
    \caption{MuRIL accuracy scores. "(not in PT)" are languages absent from MuRIL’s pretraining.}
    \label{fig:muril_accuracy}
\end{figure}

\paragraph{Zero-shot Transfer from Unknown Languages}
\label{subsec:muril results}

While the correlation analysis provides useful insights, it remains slightly inconclusive. A key limitation is that we cannot pretrain models to directly manipulate this variable for a controlled experiment. We also lack sufficient fine-tuning data for languages that were less prominent during pretraining, making it difficult to determine whether poor model performance stems from limited fine-tuning or limited exposure during pretraining. 

To counter this, we analyzed transfer in an edge case scenario. We used MuRIL (\S\ref{subsect:models}), a model pretrained exclusively on 16 Indian languages and English. Evaluating zero-shot transfer performance of models that are fine-tuned on languages entirely absent from MuRIL’s pretraining corpus can offer insight on the role of pretraining size in transfer.

Figure \ref{fig:muril_accuracy} shows graded transfer effects associated with the pretraining size, supporting the correlation analysis above. Hindi, being the dominant language in MuRIL (despite less data than English, it benefits from exposure to 16 Indic languages, including 2–4 with the same script and 9 with lexical overlap), enjoys the largest zero-shot transfer regardless of the fine-tuning language (tallest blue bars across all different models, not considering the full-shot condition in white), followed by English (yellow bars). Second, we observe substantial transfer from French, German and even Chinese on Hindi, languages MuRIL was not pretrained on (see \S\ref{sec:dicussion} for further discussion). Third, MuRIL was able to learn and perform well on German, despite it being absent from its pretraining stage. 

We find that the \textsc{multi} model, despite being trained predominantly on German, French, English, and Chinese, failed to learn any meaningful sense disambiguation, let alone transfer them to Hindi. This highlights a stark contrast: while MuRIL trained on monolingual data can effectively learn the task from languages it was not pretrained on, the multilingual data appears too noisy for effective transfer in this extreme zero-shot setting.


\subsection{Lexical Semantic Change Detection}
\label{sec:LSCD}

\begin{table}[h!]
\centering
\small
\resizebox{\columnwidth}{!}{
\begin{tabular}{|l||c|c||c|c|c|c|c|c|}
\hline
\raisebox{-5.5ex}{
\makecell[{{c}}]{LSCD\\Language $\downarrow$}}
& \rotatebox{270}{\textsc{xl-lexeme   }} & \rotatebox{270}{\textsc{multi}} & \rotatebox{270}{\textsc{german}} & \rotatebox{270}{\textsc{french}} & \rotatebox{270}{\textsc{english}} & \rotatebox{270}{\textsc{hindi}} & \rotatebox{270}{\textsc{chinese}} \\ \hline
English    & \underline{.757} & .703 & \underline{.737} & .681 & \textbf{.772} & .436 & .673 \\ \hline
German    & \textbf{.873} & \underline{.863} & \underline{.841} & \underline{.867} & \underline{.844} & .635 & .641 \\ \hline
Swedish    & \underline{.754} & \textbf{.801} & \underline{.754} & .618 & \underline{.724} & .480 & .489 \\ \hline
Latin    & -.035 & \underline{.117} & \textbf{.161} & \underline{.136} & \underline{.135} & -.177 & .091 \\ \hline
Spanish    & \underline{.665} & \underline{.696} & \underline{.670} & \underline{.664} & \textbf{.711} & .354 & .383 \\ \hline
Chinese    & \underline{.734} & .652 & .649 & .499 & \textbf{.737} & .524 & .593 \\ \hline
Norwegian$_1$ & .668 & \underline{.729} & .638 & .697 & \textbf{.777} & .525 & .400 \\ \hline
Norwegian$_2$ & \underline{.634} & \textbf{.655} & .604 & .580 & \underline{.645} & .433 & .439 \\ \hline
Average & \underline{.631} & \underline{.652} & \underline{.632} & .593 & \textbf{.668} & .401 & .464 \\ \hline
\end{tabular}}
\caption{Spearman correlations of XLM-R models' APD scores with graded semantic change scores across LSCD tasks in different languages. Best scores are in bold, scores within 0.05 of the best are underlined.}
\label{tab:lscd}
\end{table}

\footnotetext{For Norwegian, two corpora pairs are available -- the period 1929–1965 paired with 1970–2013, reported as NO$_1$, as well as 1980–1990 with 2012–2019, reported as NO$_2$.}

We evaluate XL-LEXEME \citep{cassotti2023xl}, trained on multilingual and inter-lingual data (all WiC datasets plus AM$^2$iCO), against the same \textsc{mono} models from earlier analyses. As in WiC, \textsc{multi} is matched or outperformed in all but two languages by \textsc{mono} English or German. XL-LEXEME outperforms only on German, but all well-trained \textsc{mono} models perform well on this task. Notably, \textsc{mono} English outperforms all other models on average and in 4 out of 8 test languages.
Results correlate strongly with pretraining sizes in all three models (Figure~\ref{fig:pt-corr}), echoing WiC patterns.

Zero-shot transfer performs on par with full-shot. On German, \textsc{mono} \textsc{French} and \textsc{English} outperform the \textsc{German} model, and \textsc{English} is most effective on Spanish, Chinese, and Norwegian. PRT confirms these trends, with \textsc{English} showing the highest correlation. Across mBERT and BLOOM, \textsc{mono} models often match or outperform \textsc{multi} models (Tables~\ref{tab:XLMR-lscd}-\ref{tab:bloom-lscd}).\footnote{LLaMA models were not evaluated on LSCD as prompting generative models with hundreds of usages to produce a scalar ‘change score’ is not practically feasible.}

Overall, our results show that the best results in LSC are obtained by monolingual models, dismissing multilinguality as an important factor in cross-lingual transfer also on this task.

\section{Discussion}
\label{sec:dicussion}

Disentangling the effects of multilinguality from confounding factors such as data quality and pretraining exposure is inherently challenging. Rather than providing a single definitive explanation, we identify consistent trends and highlight effects that reflect the nuanced nature of cross-lingual transfer.

\paragraph{Quantity over Multilinguality}
Our results show that the amount of fine-tuning data matters more than the number of languages included during training, as \textsc{full f.t. mono} models trained on more data typically outperform the \textsc{fixed f.t. multi} model. In both fine-tuning conditions, the \textsc{multi} model only partially outperforms monolingual models, with the \textsc{english} model often matching or exceeding it, even in the \textsc{full f.t.} setting. Taken together, this pattern of results challenges the assumption that multilingual training inherently improves zero-shot transfer; rather, data quantity drives most gains. Notably, \citet{berend-2022-combating}, who also tested the role of multilinguality in word sense disambiguation but in the pretraining stage, reached the same conclusions. Our results not only complement those of \citeauthor{berend-2022-combating} but are also more practical as multilingual models are in much wider use than monolingual ones in transfer scenarios, where the latter are actually lacking for many languages. 

\paragraph{Quality Beyond Quantity}
The frequent outperformance of the \textsc{mono} English model on most tasks could point to data quality as an important factor. Certainly, the quality of training data can impact performance, and consequently, transfer to other languages. English may enjoy datasets of higher quality, often attributed to 
their curation, annotation protocols, sense granularity and text quality \cite{philippy-etal-2023-towards}.
However, assessing or normalizing data quality across languages for polysemy disambiguation is nearly impossible without costly native-speaker annotations. Thus, this factor remains difficult to study directly. 

\paragraph{Other Underlying Factors}
\textsc{english}’s strong performance may be tied to its dominance in pretraining data which is supported by its strong overall correlations reported in \S\ref{sec:pretraining_size}. The contrasting results from MuRIL, where transfer was strongest to Hindi when Hindi had a pretraining size advantage over English, further underscore this point and provide an informative exception.

Our results show that both pretraining size and language similarity are tied to transfer, but the former has a stronger correlation. The influence of pretraining size on cross-lingual transfer is well established \cite{lauscher2020zero, srinivasan2021predicting, ahuja-etal-2023-mega}, as the correlation between language similarity and transfer performance \cite{wu2019beto, pires2019multilingual, K2020Cross-Lingual}. However, while both variables clearly influence transfer, the effect seems to depend on the specific task on which transfer is evaluated: pretraining size correlates more with semantic tasks (e.g., NLI, QA), while linguistic similarity benefits syntactic tasks (e.g., POS tagging, dependency parsing) \cite{lauscher2020zero, philippy-etal-2023-towards}. 

We suggest that sense-aware tasks, such as polysemy disambiguation and semantic change, should follow the semantic route. Given the language-specific nature of word senses, syntactic similarity offers limited benefit in the transfer of knowledge. Our findings confirm this: transfer performance in these tasks depends more on pretraining size than on linguistic similarity. The rationale of this interpretation is rooted in the type and relevance of linguistic \textit{knowledge} that is being transferred \cite{rajaee-monz-2024-analyzing, goldman2025eclektic}. We propose that syntactic tasks rely more heavily on structural similarities across languages, and thus benefit from linguistic similarity. As syntactic patterns diverge across typologically distant languages, their utility in transfer diminishes. In contrast, semantic tasks are more dependent on the diversity and scale of pretraining data, which exposes models to a wider range of meaning representations.

\paragraph{Lexical Semantic Change}
The results on LSCD align with the conclusions observed in the WiC tasks: multilinguality does not necessarily lead to improved performance, and monolingual models, particularly \textsc{english}, often outperform multilingual ones. This holds even when the test language is included in the multilingual setup (and, of course, absent from the monolingual model), as seen in the cases of Spanish, Chinese, and Norwegian, where the English model outperformed both multilingual models.
These findings reinforce the paper's central claim: multilinguality is neither necessary nor inherently beneficial for effective transfer, and monolingual models can surpass multilingual ones even under comparison conditions that favor the latter. In such cases, the multilingual model benefits from full-shot exposure to the test language, while the monolingual model is evaluated in a zero-shot setting—yet still performs better. These results further emphasize the role of dataset quality in transfer, suggesting that the perceived advantages of multilinguality may be confounded by differences in training data quality.

As lexical semantic change gains popularity and more languages are being studied using computational methods, it is important not to perpetuate the misconception that multilingual models are always the best solution—not even when the target language is included in the multilingual training setup. This conclusion is further supported by \citet{baes-etal-2025-lsc}, who recently showed that the state-of-the-art model for LSCD—the multilingual XL-LEXEME by \citet{cassotti2023xl} also included in our comparisons—is not the best choice across different semantic change scenarios. Their findings call for a more nuanced approach to model selection—one that accounts for the specific conditions of the linguistic inquiry being studied, whether it is a different language or type of change.




\paragraph{Large Models and Architectures}
Notably, larger models like LLaMA, pretrained on much more data, underperform compared to smaller models like XLM-R. This corroborates our main finding that increasing the amount of training data alone  -- whether in terms of language coverage or volume -- does not guarantee better transfer. However, architectural differences (encoder vs. decoder), hyperparameter tuning, and adaptation methods likely also influence performance and require further investigation. Additionally, \citet{berend-2022-combating} suggests that multilingual models may not even be necessary for effective transfer, highlighting that monolingual pretrained models can achieve strong cross-lingual generalization on sense-aware tasks when combined with appropriate adaptation techniques.

\section{Conclusion}

This study is the first to directly investigate the role of multilinguality in transfer for sense-aware tasks, polysemy and lexical semantic change. Our results indicate that the improved performance typically attributed to multilinguality largely stems from confounding factors such as training size, rather than linguistic diversity. Indeed, training on all available languages increases data quantity, inadvertently benefiting transfer, but this effect should not be mistakenly credited to multilingual diversity itself.

While further research is needed to determine whether these patterns generalize across other tasks, they may underline broader implications for multilingual NLP research that traditionally favored multilingual quantity over data quality. As expanding datasets across multiple languages is costly, and sometimes not practical for some tasks and languages, future research should prioritize understanding the characteristics of training regimes that facilitate transfer in order to optimize training resources for effective cross-lingual transfer. 

\section{Limitations} 

Zero-shot transfer enables models to perform tasks in languages without task-specific training data. However, its effectiveness depends on the pretrained model’s ability to capture the linguistic properties of both the source and target languages, as well as the availability of sufficient, high-quality data in the fine-tuning language.
Our study is constrained by this limitation, as we were only able to train models on a limited set of languages with highly imbalanced data distributions. As in broader cross-lingual transfer research, high-resource languages dominate, while lower-resource languages often lack sufficient data for effective fine-tuning.

We evaluated zero-shot transfer on two tasks: WiC and LSCD. Both use the same bi-encoder architecture and cosine similarity to produce their outputs. While these results provide useful insights, further research is needed to validate the findings across other tasks and model architectures.


Our evaluation of LLaMA is limited to the WiC task, and several factors constrain its interpretability. First, its pretraining data is unknown so we could not assess the correlation of performance with pretraining size. Second, LLaMA's pretraining could include the datasets used in our training or evaluation. And lastly, adapting WiC to LLaMA’s generative format (see Appendix~\ref{sec:llama_details}) introduced further limitations, such as the lack of a threshold to control sense discrimination, and a cue about the task for LLaMA, which encoder-based models lacked. Considering these, LLaMA is not directly comparable to the other models. Future work is needed to better understand the impact of these architectural differences. Due to computational resource constraints, we were unable to include other larger-scale models (8B+ parameters) or mixture-of-experts (MoE) variants in our experiments.
Despite this, our study remains broadly relevant as encoder-based models remain crucial for tasks like classification, regression, and ranking in retrieval systems, where their efficiency and lower resource demands make them well-suited. Additionally, on-device performance requires fast inference and low memory usage. While recent hardware and quantization techniques enable some larger models to run locally on powerful devices, encoder models remain preferred for many real-world applications that require fast, cost-effective, and scalable solutions, especially in specialized or low-resource settings.

Further investigation into the factors influencing zero-shot performance, particularly in low-resource languages, is essential for improving cross-lingual transfer and democratizing access to language technologies.

\section{Ethical Considerations}

This study investigates factors influencing successful zero-shot transfer, with the goal of informing the NLP community on how to develop more effective methods for speakers of low-resource languages. We evaluate models on tasks in a total of 28 languages and provide details on the languages and dataset sources used to create the datasets.

We do not foresee any direct ethical risks arising from our findings. Rather, this work promotes more responsible resource allocation by encouraging a shift away from continual dataset creation in favor of improving cross-lingual transfer techniques and understanding what defines high-quality datasets. However, it is important to acknowledge that zero-shot methods, while beneficial, may still introduce biases due to disparities in pretraining data, potentially disadvantaging underrepresented languages. Additionally, reliance on transfer from high-resource languages may reinforce linguistic hierarchies, where certain languages disproportionately influence model behavior.

Future work should continue to critically assess the impact of cross-lingual transfer on linguistic diversity and ensure that improvements in NLP benefit a wide range of language communities equitably.

\section{Acknowledgments}
We are grateful to Francesco Periti and Dominik Schlechtweg for their invaluable feedback and insightful advice on this manuscript, and to Mahmud Akhter and Pierluigi Cassotti for their support.

This work has in part been funded by the research program Change is Key! supported by Riksbankens Jubileumsfond (under reference number M21-0021).


\bibliography{custom}

\begin{thebibliography}{52}
\providecommand{\natexlab}[1]{#1}

\bibitem[{Ahuja et~al.(2023)Ahuja, Diddee, Hada, Ochieng, Ramesh, Jain, Nambi, Ganu, Segal, Ahmed, Bali, and Sitaram}]{ahuja-etal-2023-mega}
Kabir Ahuja, Harshita Diddee, Rishav Hada, Millicent Ochieng, Krithika Ramesh, Prachi Jain, Akshay Nambi, Tanuja Ganu, Sameer Segal, Mohamed Ahmed, Kalika Bali, and Sunayana Sitaram. 2023.
\newblock \href {https://doi.org/10.18653/v1/2023.emnlp-main.258} {{MEGA}: Multilingual evaluation of generative {AI}}.
\newblock In \emph{Proceedings of the 2023 Conference on Empirical Methods in Natural Language Processing}, pages 4232--4267, Singapore. Association for Computational Linguistics.

\bibitem[{AI@Meta(2024)}]{llama3modelcard}
AI@Meta. 2024.
\newblock \href {https://github.com/meta-llama/llama3/blob/main/MODEL_CARD.md} {Llama 3 model card}.

\bibitem[{Arefyev et~al.(2021)Arefyev, Homskiy, Fedoseev, Davletov, Protasov, and Panchenko}]{arefyev-2021-etal}
Nikolay Arefyev, Daniil Homskiy, Maksim Fedoseev, Adis Davletov, Vitaly Protasov, and Alexander Panchenko. 2021.
\newblock Deepmistake: Which senses are hard to distinguish for a wordincontext model.
\newblock In \emph{Computational Linguistics and Intellectual Technologies - Papers from the Annual International Conference 'Dialogue' 2021}.

\bibitem[{Baes et~al.(2025)Baes, Merx, Haslam, Vylomova, and Dubossarsky}]{baes-etal-2025-lsc}
Naomi Baes, Raphael Merx, Nick Haslam, Ekaterina Vylomova, and Haim Dubossarsky. 2025.
\newblock \href {https://doi.org/10.18653/v1/2025.findings-acl.570} {{LSC}-eval: A general framework to evaluate methods for assessing dimensions of lexical semantic change using {LLM}-generated synthetic data}.
\newblock In \emph{Findings of the Association for Computational Linguistics: ACL 2025}, pages 10905--10939, Vienna, Austria. Association for Computational Linguistics.

\bibitem[{Berend(2022)}]{berend-2022-combating}
G{\'a}bor Berend. 2022.
\newblock \href {https://doi.org/10.18653/v1/2022.naacl-main.176} {Combating the curse of multilinguality in cross-lingual {WSD} by aligning sparse contextualized word representations}.
\newblock In \emph{Proceedings of the 2022 Conference of the North American Chapter of the Association for Computational Linguistics: Human Language Technologies}, pages 2459--2471, Seattle, United States. Association for Computational Linguistics.

\bibitem[{Bevilacqua et~al.(2021)Bevilacqua, Pasini, Raganato, and Navigli}]{bevilacqua2021recent}
Michele Bevilacqua, Tommaso Pasini, Alessandro Raganato, and Roberto Navigli. 2021.
\newblock Recent trends in word sense disambiguation: A survey.
\newblock In \emph{International Joint Conference on Artificial Intelligence}, pages 4330--4338. International Joint Conference on Artificial Intelligence, Inc.

\bibitem[{Cassotti et~al.(2023)Cassotti, Siciliani, DeGemmis, Semeraro, and Basile}]{cassotti2023xl}
Pierluigi Cassotti, Lucia Siciliani, Marco DeGemmis, Giovanni Semeraro, and Pierpaolo Basile. 2023.
\newblock Xl-lexeme: Wic pretrained model for cross-lingual lexical semantic change.
\newblock In \emph{Proceedings of the 61st Annual Meeting of the Association for Computational Linguistics (Volume 2: Short Papers)}, pages 1577--1585.

\bibitem[{Chang et~al.(2024)Chang, Arnett, Tu, and Bergen}]{chang-etal-2024-multilinguality}
Tyler~A. Chang, Catherine Arnett, Zhuowen Tu, and Ben Bergen. 2024.
\newblock \href {https://doi.org/10.18653/v1/2024.emnlp-main.236} {When is multilinguality a curse? language modeling for 250 high- and low-resource languages}.
\newblock In \emph{Proceedings of the 2024 Conference on Empirical Methods in Natural Language Processing}, pages 4074--4096, Miami, Florida, USA. Association for Computational Linguistics.

\bibitem[{Chen et~al.(2023)Chen, Chersoni, Schlechtweg, Proki{\'c}, and Huang}]{chen2023chiwug}
Jing Chen, Emmanuele Chersoni, Dominik Schlechtweg, Jelena Proki{\'c}, and Chu-Ren Huang. 2023.
\newblock Chiwug: A graph-based evaluation dataset for chinese lexical semantic change detection.
\newblock In \emph{Proceedings of the 4th Workshop on Computational Approaches to Historical Language Change}, pages 93--99.

\bibitem[{Choenni et~al.(2023{\natexlab{a}})Choenni, Garrette, and Shutova}]{choenni-etal-2023-cross}
Rochelle Choenni, Dan Garrette, and Ekaterina Shutova. 2023{\natexlab{a}}.
\newblock \href {https://doi.org/10.1162/coli_a_00482} {Cross-lingual transfer with language-specific subnetworks for low-resource dependency parsing}.
\newblock \emph{Computational Linguistics}, pages 613--641.

\bibitem[{Choenni et~al.(2023{\natexlab{b}})Choenni, Garrette, and Shutova}]{choenni-etal-2023-languages}
Rochelle Choenni, Dan Garrette, and Ekaterina Shutova. 2023{\natexlab{b}}.
\newblock \href {https://doi.org/10.18653/v1/2023.emnlp-main.818} {How do languages influence each other? studying cross-lingual data sharing during {LM} fine-tuning}.
\newblock In \emph{Proceedings of the 2023 Conference on Empirical Methods in Natural Language Processing}, pages 13244--13257, Singapore. Association for Computational Linguistics.

\bibitem[{Conneau et~al.(2020)Conneau, Khandelwal, Goyal, Chaudhary, Wenzek, Guzm{\'a}n, Grave, Ott, Zettlemoyer, and Stoyanov}]{conneau2019unsupervised}
Alexis Conneau, Kartikay Khandelwal, Naman Goyal, Vishrav Chaudhary, Guillaume Wenzek, Francisco Guzm{\'a}n, Edouard Grave, Myle Ott, Luke Zettlemoyer, and Veselin Stoyanov. 2020.
\newblock \href {https://doi.org/10.18653/v1/2020.acl-main.747} {Unsupervised cross-lingual representation learning at scale}.
\newblock In \emph{Proceedings of the 58th Annual Meeting of the Association for Computational Linguistics}, pages 8440--8451, Online. Association for Computational Linguistics.

\bibitem[{Dairkee and Dubossarsky(2024)}]{dubossarsky2024strengthening}
Farheen Dairkee and Haim Dubossarsky. 2024.
\newblock Strengthening the wic: New polysemy dataset in hindi and lack of cross lingual transfer.
\newblock In \emph{Proceedings of the 2024 Joint International Conference on Computational Linguistics, Language Resources and Evaluation (LREC-COLING 2024)}, pages 15341--15349.

\bibitem[{de~Vries et~al.(2022)de~Vries, Wieling, and Nissim}]{de-vries-etal-2022-make}
Wietse de~Vries, Martijn Wieling, and Malvina Nissim. 2022.
\newblock \href {https://doi.org/10.18653/v1/2022.acl-long.529} {Make the best of cross-lingual transfer: Evidence from {POS} tagging with over 100 languages}.
\newblock In \emph{Proceedings of the 60th Annual Meeting of the Association for Computational Linguistics (Volume 1: Long Papers)}, pages 7676--7685, Dublin, Ireland. Association for Computational Linguistics.

\bibitem[{Devlin et~al.(2019)Devlin, Chang, Lee, and Toutanova}]{devlin2018bert}
Jacob Devlin, Ming-Wei Chang, Kenton Lee, and Kristina Toutanova. 2019.
\newblock \href {https://doi.org/10.18653/v1/N19-1423} {{BERT}: Pre-training of deep bidirectional transformers for language understanding}.
\newblock In \emph{Proceedings of the 2019 Conference of the North {A}merican Chapter of the Association for Computational Linguistics: Human Language Technologies, Volume 1 (Long and Short Papers)}, pages 4171--4186, Minneapolis, Minnesota. Association for Computational Linguistics.

\bibitem[{Dolicki and Spanakis(2021)}]{dolicki2021analysing}
B{\l}a{\.z}ej Dolicki and Gerasimos Spanakis. 2021.
\newblock Analysing the impact of linguistic features on cross-lingual transfer.
\newblock \emph{arXiv preprint arXiv:2105.05975}.

\bibitem[{Ebrahimi and Kann(2021)}]{ebrahimi-kann-2021-adapt}
Abteen Ebrahimi and Katharina Kann. 2021.
\newblock \href {https://doi.org/10.18653/v1/2021.acl-long.351} {How to adapt your pretrained multilingual model to 1600 languages}.
\newblock In \emph{Proceedings of the 59th Annual Meeting of the Association for Computational Linguistics and the 11th International Joint Conference on Natural Language Processing (Volume 1: Long Papers)}, pages 4555--4567, Online. Association for Computational Linguistics.

\bibitem[{Goldman et~al.(2025)Goldman, Shaham, Malkin, Eiger, Hassidim, Matias, Maynez, Gilady, Riesa, Rijhwani et~al.}]{goldman2025eclektic}
Omer Goldman, Uri Shaham, Dan Malkin, Sivan Eiger, Avinatan Hassidim, Yossi Matias, Joshua Maynez, Adi~Mayrav Gilady, Jason Riesa, Shruti Rijhwani, et~al. 2025.
\newblock Eclektic: a novel challenge set for evaluation of cross-lingual knowledge transfer.
\newblock \emph{arXiv preprint arXiv:2502.21228}.

\bibitem[{Goworek et~al.(2025)Goworek, Karlcut, Shezad, Darshana, Mane, Bondada, Sikka, Mammadov, Allahverdiyev, Purighella, Gupta, Ndegwa, Tran, and Dubossarsky}]{goworek-etal-2025-senwich}
Roksana Goworek, Harpal~Singh Karlcut, Hamza Shezad, Nijaguna Darshana, Abhishek Mane, Syam Bondada, Raghav Sikka, Ulvi Mammadov, Rauf Allahverdiyev, Sriram~Satkirti Purighella, Paridhi Gupta, Muhinyia Ndegwa, Bao~Khanh Tran, and Haim Dubossarsky. 2025.
\newblock \href {https://doi.org/10.18653/v1/2025.sigtyp-1.7} {{S}en{W}i{C}h: Sense-annotation of low-resource languages for {W}i{C} using hybrid methods}.
\newblock In \emph{Proceedings of the 7th Workshop on Research in Computational Linguistic Typology and Multilingual NLP}, pages 61--74, Vienna, Austria. Association for Computational Linguistics.

\bibitem[{Grattafiori et~al.(2024)Grattafiori, Dubey, Jauhri, Pandey, Kadian, Al-Dahle, Letman, Mathur, Schelten, Vaughan et~al.}]{grattafiori2024llama}
Aaron Grattafiori, Abhimanyu Dubey, Abhinav Jauhri, Abhinav Pandey, Abhishek Kadian, Ahmad Al-Dahle, Aiesha Letman, Akhil Mathur, Alan Schelten, Alex Vaughan, et~al. 2024.
\newblock The llama 3 herd of models.
\newblock \emph{arXiv preprint arXiv:2407.21783}.

\bibitem[{Imani et~al.(2023)Imani, Lin, Kargaran, Severini, Jalili~Sabet, Kassner, Ma, Schmid, Martins, Yvon, and Sch{\"u}tze}]{imanigooghari-etal-2023-glot500}
Ayyoob Imani, Peiqin Lin, Amir~Hossein Kargaran, Silvia Severini, Masoud Jalili~Sabet, Nora Kassner, Chunlan Ma, Helmut Schmid, Andr{\'e} Martins, Fran{\c{c}}ois Yvon, and Hinrich Sch{\"u}tze. 2023.
\newblock \href {https://doi.org/10.18653/v1/2023.acl-long.61} {Glot500: Scaling multilingual corpora and language models to 500 languages}.
\newblock In \emph{Proceedings of the 61st Annual Meeting of the Association for Computational Linguistics (Volume 1: Long Papers)}, pages 1082--1117, Toronto, Canada. Association for Computational Linguistics.

\bibitem[{K et~al.(2020)K, Wang, Mayhew, and Roth}]{K2020Cross-Lingual}
Karthikeyan K, Zihan Wang, Stephen Mayhew, and Dan Roth. 2020.
\newblock \href {https://openreview.net/forum?id=HJeT3yrtDr} {Cross-lingual ability of multilingual bert: An empirical study}.
\newblock In \emph{International Conference on Learning Representations}.

\bibitem[{Khanuja et~al.(2021)Khanuja, Bansal, Mehtani, Khosla, Dey, Gopalan, Margam, Aggarwal, Nagipogu, Dave, Gupta, Gali, Subramanian, and Talukdar}]{khanuja2021muril}
Simran Khanuja, Diksha Bansal, Sarvesh Mehtani, Savya Khosla, Atreyee Dey, Balaji Gopalan, Dilip~Kumar Margam, Pooja Aggarwal, Rajiv~Teja Nagipogu, Shachi Dave, Shruti Gupta, Subhash Chandra~Bose Gali, Vish Subramanian, and Partha Talukdar. 2021.
\newblock \href {https://arxiv.org/abs/2103.10730} {Muril: Multilingual representations for indian languages}.
\newblock \emph{Preprint}, arXiv:2103.10730.

\bibitem[{Kutuzov and Giulianelli(2020)}]{kutuzov2020uio}
Andrey Kutuzov and Mario Giulianelli. 2020.
\newblock \href {https://doi.org/10.18653/v1/2020.semeval-1.14} {{U}i{O}-{U}v{A} at {S}em{E}val-2020 task 1: Contextualised embeddings for lexical semantic change detection}.
\newblock In \emph{Proceedings of the Fourteenth Workshop on Semantic Evaluation}, pages 126--134, Barcelona (online). International Committee for Computational Linguistics.

\bibitem[{Kutuzov et~al.(2022)Kutuzov, Touileb, M{\ae}hlum, Enstad, and Wittemann}]{kutuzov2022nordiachange}
Andrey Kutuzov, Samia Touileb, Petter M{\ae}hlum, Tita Enstad, and Alexandra Wittemann. 2022.
\newblock \href {https://aclanthology.org/2022.lrec-1.274/} {{N}or{D}ia{C}hange: Diachronic semantic change dataset for {N}orwegian}.
\newblock In \emph{Proceedings of the Thirteenth Language Resources and Evaluation Conference}, pages 2563--2572, Marseille, France. European Language Resources Association.

\bibitem[{Lauscher et~al.(2020)Lauscher, Ravishankar, Vuli{\'c}, and Glava{\v{s}}}]{lauscher2020zero}
Anne Lauscher, Vinit Ravishankar, Ivan Vuli{\'c}, and Goran Glava{\v{s}}. 2020.
\newblock \href {https://doi.org/10.18653/v1/2020.emnlp-main.363} {From zero to hero: {O}n the limitations of zero-shot language transfer with multilingual {T}ransformers}.
\newblock In \emph{Proceedings of the 2020 Conference on Empirical Methods in Natural Language Processing (EMNLP)}, pages 4483--4499, Online. Association for Computational Linguistics.

\bibitem[{Le~Scao et~al.(2023)Le~Scao, Fan, Akiki, Pavlick, Ili{\'c}, Hesslow, Castagn{\'e}, Luccioni, Yvon, Gall{\'e} et~al.}]{le2023bloom}
Teven Le~Scao, Angela Fan, Christopher Akiki, Ellie Pavlick, Suzana Ili{\'c}, Daniel Hesslow, Roman Castagn{\'e}, Alexandra~Sasha Luccioni, Fran{\c{c}}ois Yvon, Matthias Gall{\'e}, et~al. 2023.
\newblock Bloom: A 176b-parameter open-access multilingual language model.

\bibitem[{Littell et~al.(2017)Littell, Mortensen, Lin, Kairis, Turner, and Levin}]{littell-etal-2017-uriel}
Patrick Littell, David~R. Mortensen, Ke~Lin, Katherine Kairis, Carlisle Turner, and Lori Levin. 2017.
\newblock \href {https://aclanthology.org/E17-2002/} {{URIEL} and lang2vec: Representing languages as typological, geographical, and phylogenetic vectors}.
\newblock In \emph{Proceedings of the 15th Conference of the {E}uropean Chapter of the Association for Computational Linguistics: Volume 2, Short Papers}, pages 8--14, Valencia, Spain. Association for Computational Linguistics.

\bibitem[{Liu et~al.(2021)Liu, Ponti, McCarthy, Vuli{\'c}, and Korhonen}]{liu2021am2ico}
Qianchu Liu, Edoardo~Maria Ponti, Diana McCarthy, Ivan Vuli{\'c}, and Anna Korhonen. 2021.
\newblock \href {https://doi.org/10.18653/v1/2021.emnlp-main.571} {{AM}2i{C}o: Evaluating word meaning in context across low-resource languages with adversarial examples}.
\newblock In \emph{Proceedings of the 2021 Conference on Empirical Methods in Natural Language Processing}, pages 7151--7162, Online and Punta Cana, Dominican Republic. Association for Computational Linguistics.

\bibitem[{Martelli et~al.(2021)Martelli, Kalach, Tola, Navigli et~al.}]{martelli2021semeval}
Federico Martelli, Najla Kalach, Gabriele Tola, Roberto Navigli, et~al. 2021.
\newblock Semeval-2021 task 2: Multilingual and cross-lingual word-in-context disambiguation (mcl-wic).
\newblock In \emph{Proceedings of the 15th International Workshop on Semantic Evaluation (SemEval-2021)}, pages 24--36.

\bibitem[{Navigli(2009)}]{navigli2009word}
Roberto Navigli. 2009.
\newblock Word sense disambiguation: A survey.
\newblock \emph{ACM computing surveys (CSUR)}, 41(2):1--69.

\bibitem[{Navigli and Ponzetto(2010)}]{navigli-ponzetto-2010-babelnet}
Roberto Navigli and Simone~Paolo Ponzetto. 2010.
\newblock \href {https://aclanthology.org/P10-1023/} {{B}abel{N}et: Building a very large multilingual semantic network}.
\newblock In \emph{Proceedings of the 48th Annual Meeting of the Association for Computational Linguistics}, pages 216--225, Uppsala, Sweden. Association for Computational Linguistics.

\bibitem[{Patil et~al.(2022)Patil, Talukdar, and Sarawagi}]{patil-etal-2022-overlap}
Vaidehi Patil, Partha Talukdar, and Sunita Sarawagi. 2022.
\newblock \href {https://doi.org/10.18653/v1/2022.acl-long.18} {Overlap-based vocabulary generation improves cross-lingual transfer among related languages}.
\newblock In \emph{Proceedings of the 60th Annual Meeting of the Association for Computational Linguistics (Volume 1: Long Papers)}, pages 219--233, Dublin, Ireland. Association for Computational Linguistics.

\bibitem[{Periti and Montanelli(2024)}]{periti2024lexical}
Francesco Periti and Stefano Montanelli. 2024.
\newblock \href {https://doi.org/10.1145/3672393} {Lexical semantic change through large language models: a survey}.
\newblock \emph{ACM Comput. Surv.}, 56(11).

\bibitem[{Periti and Tahmasebi(2024)}]{periti2024systematic}
Francesco Periti and Nina Tahmasebi. 2024.
\newblock \href {https://doi.org/10.18653/v1/2024.naacl-long.240} {A systematic comparison of contextualized word embeddings for lexical semantic change}.
\newblock In \emph{Proceedings of the 2024 Conference of the North American Chapter of the Association for Computational Linguistics: Human Language Technologies (Volume 1: Long Papers)}, pages 4262--4282, Mexico City, Mexico. Association for Computational Linguistics.

\bibitem[{Philippy et~al.(2023)Philippy, Guo, and Haddadan}]{philippy-etal-2023-towards}
Fred Philippy, Siwen Guo, and Shohreh Haddadan. 2023.
\newblock \href {https://doi.org/10.18653/v1/2023.acl-long.323} {Towards a common understanding of contributing factors for cross-lingual transfer in multilingual language models: A review}.
\newblock In \emph{Proceedings of the 61st Annual Meeting of the Association for Computational Linguistics (Volume 1: Long Papers)}, pages 5877--5891, Toronto, Canada. Association for Computational Linguistics.

\bibitem[{Pilehvar and Camacho-Collados(2019)}]{pilehvar2018wic}
Mohammad~Taher Pilehvar and Jose Camacho-Collados. 2019.
\newblock \href {https://doi.org/10.18653/v1/N19-1128} {{W}i{C}: the word-in-context dataset for evaluating context-sensitive meaning representations}.
\newblock In \emph{Proceedings of the 2019 Conference of the North {A}merican Chapter of the Association for Computational Linguistics: Human Language Technologies, Volume 1 (Long and Short Papers)}, pages 1267--1273, Minneapolis, Minnesota. Association for Computational Linguistics.

\bibitem[{Pires et~al.(2019)Pires, Schlinger, and Garrette}]{pires2019multilingual}
Telmo Pires, Eva Schlinger, and Dan Garrette. 2019.
\newblock \href {https://doi.org/10.18653/v1/P19-1493} {How multilingual is multilingual {BERT}?}
\newblock In \emph{Proceedings of the 57th Annual Meeting of the Association for Computational Linguistics}, pages 4996--5001, Florence, Italy. Association for Computational Linguistics.

\bibitem[{Ponti et~al.(2018)Ponti, Vuli{\'c}, Glava{\v{s}}, Mrk{\v{s}}i{\'c}, and Korhonen}]{ponti2018adversarial}
Edoardo~Maria Ponti, Ivan Vuli{\'c}, Goran Glava{\v{s}}, Nikola Mrk{\v{s}}i{\'c}, and Anna Korhonen. 2018.
\newblock \href {https://doi.org/10.18653/v1/D18-1026} {Adversarial propagation and zero-shot cross-lingual transfer of word vector specialization}.
\newblock In \emph{Proceedings of the 2018 Conference on Empirical Methods in Natural Language Processing}, pages 282--293, Brussels, Belgium. Association for Computational Linguistics.

\bibitem[{Raganato et~al.(2020)Raganato, Pasini, Camacho-Collados, and Pilehvar}]{raganato2020xl}
Alessandro Raganato, Tommaso Pasini, Jose Camacho-Collados, and Mohammad~Taher Pilehvar. 2020.
\newblock Xl-wic: A multilingual benchmark for evaluating semantic contextualization.
\newblock In \emph{Proceedings of the 2020 Conference on Empirical Methods in Natural Language Processing (EMNLP)}, pages 7193--7206.

\bibitem[{Rajaee and Monz(2024)}]{rajaee-monz-2024-analyzing}
Sara Rajaee and Christof Monz. 2024.
\newblock \href {https://doi.org/10.18653/v1/2024.eacl-long.177} {Analyzing the evaluation of cross-lingual knowledge transfer in multilingual language models}.
\newblock In \emph{Proceedings of the 18th Conference of the European Chapter of the Association for Computational Linguistics (Volume 1: Long Papers)}, pages 2895--2914, St. Julian{'}s, Malta. Association for Computational Linguistics.

\bibitem[{Rzymski et~al.(2020)Rzymski, Tresoldi, Greenhill, Wu, Schweikhard, Koptjevskaja-Tamm, Gast, Bodt, Hantgan, Kaiping et~al.}]{rzymski2020database}
Christoph Rzymski, Tiago Tresoldi, Simon~J Greenhill, Mei-Shin Wu, Nathanael~E Schweikhard, Maria Koptjevskaja-Tamm, Volker Gast, Timotheus~A Bodt, Abbie Hantgan, Gereon~A Kaiping, et~al. 2020.
\newblock The database of cross-linguistic colexifications, reproducible analysis of cross-linguistic polysemies.
\newblock \emph{Scientific data}, 7(1):13.

\bibitem[{Sainz et~al.(2023)Sainz, Campos, Garc{\'i}a-Ferrero, Etxaniz, de~Lacalle, and Agirre}]{sainz-etal-2023-nlp}
Oscar Sainz, Jon Campos, Iker Garc{\'i}a-Ferrero, Julen Etxaniz, Oier~Lopez de~Lacalle, and Eneko Agirre. 2023.
\newblock \href {https://doi.org/10.18653/v1/2023.findings-emnlp.722} {{NLP} evaluation in trouble: On the need to measure {LLM} data contamination for each benchmark}.
\newblock In \emph{Findings of the Association for Computational Linguistics: EMNLP 2023}, pages 10776--10787, Singapore. Association for Computational Linguistics.

\bibitem[{Schlechtweg et~al.(2020)Schlechtweg, McGillivray, Hengchen, Dubossarsky, and Tahmasebi}]{schlechtweg2020semeval}
Dominik Schlechtweg, Barbara McGillivray, Simon Hengchen, Haim Dubossarsky, and Nina Tahmasebi. 2020.
\newblock \href {https://doi.org/10.18653/v1/2020.semeval-1.1} {{S}em{E}val-2020 task 1: Unsupervised lexical semantic change detection}.
\newblock In \emph{Proceedings of the Fourteenth Workshop on Semantic Evaluation}, pages 1--23, Barcelona (online). International Committee for Computational Linguistics.

\bibitem[{Shaham et~al.(2024)Shaham, Herzig, Aharoni, Szpektor, Tsarfaty, and Eyal}]{shaham2024multilingual}
Uri Shaham, Jonathan Herzig, Roee Aharoni, Idan Szpektor, Reut Tsarfaty, and Matan Eyal. 2024.
\newblock \href {https://doi.org/10.18653/v1/2024.findings-acl.136} {Multilingual instruction tuning with just a pinch of multilinguality}.
\newblock In \emph{Findings of the Association for Computational Linguistics: ACL 2024}, pages 2304--2317, Bangkok, Thailand. Association for Computational Linguistics.

\bibitem[{Singh and Siddiqui(2016)}]{singh2016sense}
Satyendr Singh and Tanveer~J Siddiqui. 2016.
\newblock Sense annotated hindi corpus.
\newblock In \emph{2016 International Conference on Asian Language Processing (IALP)}, pages 22--25. IEEE.

\bibitem[{Srinivasan et~al.(2021)Srinivasan, Sitaram, Ganu, Dandapat, Bali, and Choudhury}]{srinivasan2021predicting}
Anirudh Srinivasan, Sunayana Sitaram, Tanuja Ganu, Sandipan Dandapat, Kalika Bali, and Monojit Choudhury. 2021.
\newblock Predicting the performance of multilingual nlp models.
\newblock \emph{arXiv preprint arXiv:2110.08875}.

\bibitem[{Wang et~al.(2023)Wang, Huang, Chang, and Chen}]{wang-etal-2023-self-augmentation}
Fei Wang, Kuan-Hao Huang, Kai-Wei Chang, and Muhao Chen. 2023.
\newblock \href {https://doi.org/10.18653/v1/2023.ijcnlp-short.1} {Self-augmentation improves zero-shot cross-lingual transfer}.
\newblock In \emph{Proceedings of the 13th International Joint Conference on Natural Language Processing and the 3rd Conference of the Asia-Pacific Chapter of the Association for Computational Linguistics (Volume 2: Short Papers)}, pages 1--9, Nusa Dua, Bali. Association for Computational Linguistics.

\bibitem[{Wu and Dredze(2019)}]{wu2019beto}
Shijie Wu and Mark Dredze. 2019.
\newblock \href {https://doi.org/10.18653/v1/D19-1077} {Beto, bentz, becas: The surprising cross-lingual effectiveness of {BERT}}.
\newblock In \emph{Proceedings of the 2019 Conference on Empirical Methods in Natural Language Processing and the 9th International Joint Conference on Natural Language Processing (EMNLP-IJCNLP)}, pages 833--844, Hong Kong, China. Association for Computational Linguistics.

\bibitem[{Wu and Dredze(2020)}]{wu2020all}
Shijie Wu and Mark Dredze. 2020.
\newblock \href {https://doi.org/10.18653/v1/2020.repl4nlp-1.16} {Are all languages created equal in multilingual {BERT}?}
\newblock In \emph{Proceedings of the 5th Workshop on Representation Learning for NLP}, pages 120--130, Online. Association for Computational Linguistics.

\bibitem[{Zamora-Reina et~al.(2022)Zamora-Reina, Bravo-Marquez, and Schlechtweg}]{zamora2022lscdiscovery}
Frank~D. Zamora-Reina, Felipe Bravo-Marquez, and Dominik Schlechtweg. 2022.
\newblock \href {https://doi.org/10.18653/v1/2022.lchange-1.16} {{LSCD}iscovery: A shared task on semantic change discovery and detection in {S}panish}.
\newblock In \emph{Proceedings of the 3rd Workshop on Computational Approaches to Historical Language Change}, pages 149--164, Dublin, Ireland. Association for Computational Linguistics.

\bibitem[{Ziemski et~al.(2016)Ziemski, Junczys-Dowmunt, and Pouliquen}]{ziemski-etal-2016-united}
Micha{\l} Ziemski, Marcin Junczys-Dowmunt, and Bruno Pouliquen. 2016.
\newblock \href {https://aclanthology.org/L16-1561/} {The {U}nited {N}ations parallel corpus v1.0}.
\newblock In \emph{Proceedings of the Tenth International Conference on Language Resources and Evaluation ({LREC}`16)}, pages 3530--3534, Portoro{\v{z}}, Slovenia. European Language Resources Association (ELRA).

\end{thebibliography}

\appendix

\newpage
\section{Model Training Details}
\label{app:hyperparameters}

\begin{table}[htbp]
    \centering
    \small
    \begin{tabular}{lll} \hline
    \toprule

        \textbf{Model Type} & \textbf{Model Name} & \textbf{Training Size} \\ 
        \midrule
         \textsc{multi} & \textsc{multi} & 122.4k \\ 
         \midrule
         \multirow{5}{*}{\textsc{mono}} & \textsc{german} & 54.7k \\ 
         & \textsc{french} & 46.1k \\ 
         & \textsc{english} & 15.1k \\
         & \textsc{hindi} & 7k \\ 
         & \textsc{chinese} & 2.5k \\ 
        \bottomrule
    \end{tabular}
    \caption{Fine-tuning sizes.}
    \label{tab:model_description}
\end{table}

\begin{figure}[htbp]
    \centering
    \includegraphics[width=\linewidth]{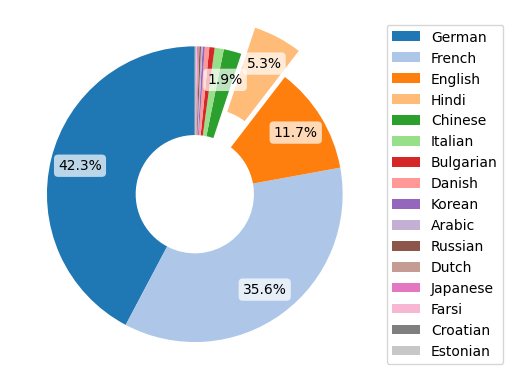}
    \caption{Proportions of languages in WiC datasets used for training. \textsc{multi} is trained on all of them except Hindi.
    }
    \label{fig:pie_chart}
\end{figure}

\begin{table}[h]
\centering
\small
\begin{tabular}{ll}
\toprule
\textbf{Model} & \textbf{Number of Parameters} \\
\midrule
XLM-R & 560M \\
BLOOM & 560M \\
mBERT & 178M \\
LLaMA3-8B-Instruct & 8B \\
MuRIL & 506M \\
\bottomrule
\end{tabular}
\caption{Number of parameters for each selected pretrained model.}
\label{tab:model-parameters}
\end{table}

\begin{table}[h]
\centering
\small
\begin{tabular}{ll}
\toprule
\textbf{Hyperparameter} & \textbf{Value} \\
\midrule
Hidden activation          & \texttt{gelu} \\
Hidden dropout probability & 0.1 \\
Hidden size                & 1024 \\
Initializer range          & 0.02 \\
Intermediate size          & 4096 \\
Layer norm epsilon         & $1 \times 10^{-5}$ \\
Max position embeddings    & 514 \\
Number of attention heads  & 16 \\
Number of hidden layers    & 24 \\
Position embedding type    & Absolute \\
Vocabulary size            & 250004 \\
Learning rate              & $1 \times 10^{-5}$ \\
Weight decay               & 0.0 \\
Max sequence length ($\lambda$) & 128 \\
\bottomrule
\end{tabular}
\caption{Fine-tuning hyperparameters used in our experiments.}
\label{tab:hyperparameters}
\end{table}

\newpage
\section{Percentage Change Between Full F.T. - Fixed F.T.}
\label{app:percentage_change}

\begin{table}[!ht]
\small
    \centering
    \begin{tabular}{|l|l|l|l|l|}
    \hline
       \diagbox{Model}{Dataset} & MCL & XL & Hindi & Am$^2$iCO \\ \hline
        \textsc{multi} & -3.3 & -5.1 & -12.3 & -2.4 \\ \hline
        \textsc{english} & 1.6 & 2.1 & 0.1 & 1.5 \\ \hline
        \textsc{german} & 1.6 & 1.5 & -18.2 & -3.0 \\ \hline
        \textsc{french} & -7.7 & 0.3 & 23.1 & -2.8 \\ \hline
    \end{tabular}
    \caption{Percentage change in average accuracy of mBERT per dataset (Fixed F.T. - Full F.T.)}
    \label{tab:mbert_ft_drop_in_perf}
\end{table}

\begin{table}[!ht]
\small
    \centering
    \begin{tabular}{|l|l|l|l|l|}
    \hline
       \diagbox{Model}{Dataset} & MCL & XL & Hindi & Am$^2$iCO \\ \hline
        \textsc{multi} & -11.5 & -9.1 & -13.0 & -4.8 \\ \hline
        \textsc{english} & -5.3 & -0.9 & -24.9 & -3.2 \\ \hline
        \textsc{german} & -11.5 & -5.8 & -20.4 & -3.7 \\ \hline
        \textsc{french} & -17.6 & -5.4 & -24.5 & -4.0 \\ \hline
    \end{tabular}
    \caption{Percentage change in average accuracy of BLOOM per dataset (Fixed F.T. - Full F.T.)}
    \label{tab:bloom_ft_drop_in_perf}
\end{table}

\begin{table}[!ht]
\small
    \centering
    \begin{tabular}{|l|l|l|l|l|}
    \hline
       \diagbox{Model}{Dataset} & MCL & XL & Hindi & Am$^2$iCO \\ \hline
        \textsc{multi} & -5.9 & -2.2 & 12.8 & 0.5 \\ \hline
        \textsc{english} & -1.9 & 3.5 & 3.6 & -2.1 \\ \hline
        \textsc{german} & -1.4 & -1.5 & -18.3 & 0.7 \\ \hline
        \textsc{french} & -1.1 & -0.8 & 4.8 & -2.6 \\ \hline
    \end{tabular}
    \caption{Percentage change in average accuracy of LLaMA per dataset (Fixed F.T. - Full F.T.)}
    \label{tab:llama_ft_drop_in_perf}
\end{table}

\newpage
\section{Linguistic Similarity}
\label{fig:corr_linguistic_similarity}
\begin{figure}[!ht]
    \centering
    \includegraphics[width=\linewidth]{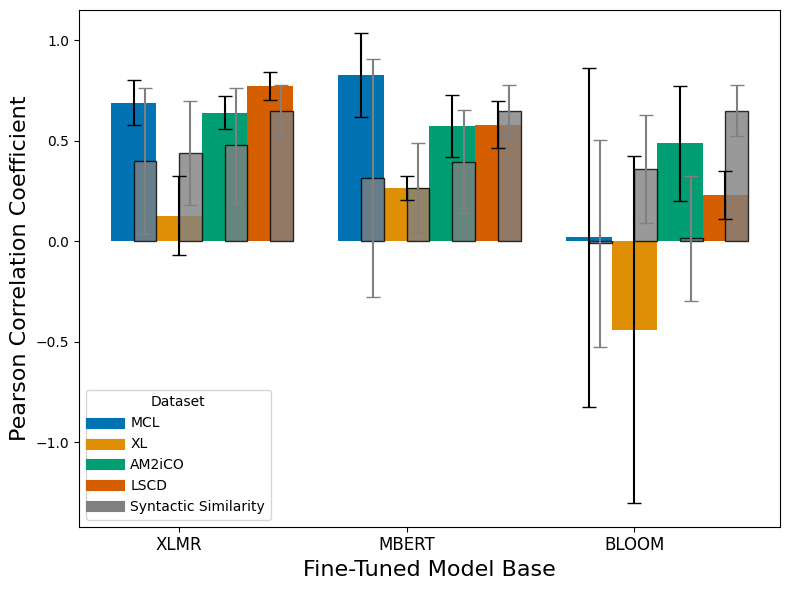}
    \caption{Mean correlations of syntactic similarity (between the fine-tuning language of \textsc{mono} models and the target language) and zero-shot performance on the target languages. Superimposed on the correlation reported in Figure~\ref{fig:pt-corr}) for comparison to correlations with pretraining sizes.}
    \label{fig:syntactic_similarity_corr}
\end{figure}

\section{Zero-Shot Comparison of mBERT and BLOOM Models}

    
    
    
    
    

    
\begin{table}[h!]
\centering
\tiny
\resizebox{\columnwidth}{!}{
\begin{tabular}{|c|c|c|c|c|c|c|c|c|}
\hline
    \raisebox{-3.5ex}{Language} & \rotatebox{270}{German } 
    & \rotatebox{270}{French} 
    & \rotatebox{270}{French} 
    & \rotatebox{270}{English} 
    & \rotatebox{270}{English} 
    & \rotatebox{270}{Hindi} 
    & \rotatebox{270}{Chinese } 
    & \rotatebox{270}{Chinese } \\ \hline
    Dataset & \rotatebox{0}{XL} 
    & \rotatebox{0}{MCL} 
    & \rotatebox{0}{XL} 
    & \rotatebox{0}{MCL} 
    & \rotatebox{0}{XL} 
    & \rotatebox{0}{Hindi} 
    & \rotatebox{0}{MCL} 
    & \rotatebox{0}{XL} \\ \hline
\textsc{z.s. multi$_{\otimes=}$} & 62.9 & 74.9 & 62.7 & 69.9 & \textbf{64.6} & 58.4 & 65.8 & 64.5 \\ \hline
\textsc{german$_=$} & \cellcolor{gray} 73.9 & 72.2 & 61.3 & \textbf{73.6} & 56.3 & 51.0 & 67.7 & 63.9 \\ \hline
\textsc{french$_=$} & \textbf{66.2} & \cellcolor{gray} 66.0 & \cellcolor{gray} 65.9 & 72.5 & 62.8 & 46.2 & 58.2 & 60.4 \\ \hline
\textsc{english$_=$} & 65.6 & \textbf{78.3} & \textbf{62.8} & \cellcolor{gray} 84.3 & \cellcolor{gray} 63.9 & \textbf{73.4} & \textbf{73.0} & \textbf{68.1} \\ \hline
\textsc{hindi$_=$} & 55.7 & 67.0 & 56.4 & 69.6 & 55.9 & \cellcolor{gray} 76.1 & 64.1 & 60.7 \\ \hline
\textsc{chinese$_=$} & 61.9 & 72.7 & 63.4 & 78.1 & 61.4 & 59.7 & \cellcolor{gray} 70.8 & \cellcolor{gray} 68.0 \\ \hline
\end{tabular}}
\caption{mBERT (Fixed F.T.) Zero-shot performance of \textsc{multi} models (\textsc{z.s. multi}) and monolingual models not trained on the target language. Grey cells indicate full-shot scenarios. All trained on 8,750 examples.}
\label{tab:zero-shot_mBERT}
\end{table}

    
    
       
    
      

    
\begin{table}[h!]
\centering
\tiny
\resizebox{\columnwidth}{!}{
\begin{tabular}{|c|c|c|c|c|c|c|c|c|}
\hline
    \raisebox{-3.5ex}{Language} & \rotatebox{270}{German } 
    & \rotatebox{270}{French} 
    & \rotatebox{270}{French} 
    & \rotatebox{270}{English} 
    & \rotatebox{270}{English} 
    & \rotatebox{270}{Hindi} 
    & \rotatebox{270}{Chinese } 
    & \rotatebox{270}{Chinese } \\ \hline
    Dataset & \rotatebox{0}{XL} 
    & \rotatebox{0}{MCL} 
    & \rotatebox{0}{XL} 
    & \rotatebox{0}{MCL} 
    & \rotatebox{0}{XL} 
    & \rotatebox{0}{Hindi} 
    & \rotatebox{0}{MCL} 
    & \rotatebox{0}{XL} \\ \hline
\textsc{z.s. multi$_{\otimes=}$} & 48.6 & 54.1 & 52.9 & \textbf{54.1} & \textbf{53.9} & \textbf{57.9} & 50.0 & 50.1 \\ \hline
\textsc{german$_=$}    &\cellcolor{gray} 61.3          & 49.4 & 50.7 & 49.7 & 51.0 & 48.4 & 50.0 & 50.0 \\ \hline
\textsc{french$_=$}    & 50.3          & \cellcolor{gray} 52.3 & \cellcolor{gray} 60.4 & 51.1 & 53.1 & 45.1 & 50.0 & 50.0 \\ \hline
\textsc{english$_=$}    & 51.2         & \textbf{67.4} & \textbf{57.1} & \cellcolor{gray} 74.6 & \cellcolor{gray} 58.6 & 48.2 & \textbf{55.6} & \textbf{52.8} \\ \hline
\textsc{hindi$_=$}   & 49.7            & 50.0 & 50.1 & 50.0 & 49.9 & \cellcolor{gray} 88.6 & 50.0 & 50.0 \\ \hline
\textsc{chinese$_=$}   & 49.8          & 50.0 & 50.1 & 50.0 & 50.0 & 46.2 & \cellcolor{gray} 54.5 & \cellcolor{gray} 59 \\ \hline
\end{tabular}}
\caption{BLOOM (Fixed F.T.) Zero-shot performance of \textsc{multi} models (\textsc{z.s. multi}) and \textsc{mono} models not trained on the target language. Grey cells indicate full-shot scenarios. All trained on 8,750 examples.}
\label{tab:zero-shot_bloom}
\end{table}


    
   

  



\begin{table}[h!]
\centering
\tiny
\resizebox{\columnwidth}{!}{
\begin{tabular}{|c|c|c|c|c|c|c|c|c|}
\hline
    \raisebox{-3.5ex}{Language} & \rotatebox{270}{German } 
    & \rotatebox{270}{French} 
    & \rotatebox{270}{French} 
    & \rotatebox{270}{English} 
    & \rotatebox{270}{English} 
    & \rotatebox{270}{Hindi} 
    & \rotatebox{270}{Chinese } 
    & \rotatebox{270}{Chinese } \\ \hline
    Dataset & \rotatebox{0}{XL} 
    & \rotatebox{0}{MCL} 
    & \rotatebox{0}{XL} 
    & \rotatebox{0}{MCL} 
    & \rotatebox{0}{XL} 
    & \rotatebox{0}{Hindi} 
    & \rotatebox{0}{MCL} 
    & \rotatebox{0}{XL} \\ \hline
\textsc{z.s. multi$_{\otimes=}$} &  66.7 & \textbf{77.6} & 67.4 & 74.1 & 68.4 & \textbf{71.5} & 72.7 & 68.9 \\ \hline
\textsc{german$_=$} &  \cellcolor{gray}76.6 & 77.1 & \textbf{68.0} & \textbf{77.5} & 64.9 & 47.2 & \textbf{75.9} & \textbf{71.9} \\ \hline
\textsc{french$_=$}& 68.6 & \cellcolor{gray} 70.5 & \cellcolor{gray} 65.8 & 77.1 & 67.8 & 64.5 & 74.5 & 71.6 \\ \hline
\textsc{english$_=$} & 63.6 & 74.2 & 64.5 & \cellcolor{gray} 84.0 & \cellcolor{gray} 70.2 & 66.2 & 71.1 & 64.4 \\ \hline
\textsc{hindi$_=$} & 59.8 & 66.4 & 56.8 & 72.7 & 64.1 & \cellcolor{gray} 72.9 & 69.4 & 64.7 \\ \hline
\textsc{chinese$_=$} & 64.0 & 70.9 & 62.1 & 70.7 & \textbf{68.7} & 68.5 & \cellcolor{gray} 71.5 & \cellcolor{gray} 70.2 \\ \hline
\end{tabular}}
\caption{LLaMA (Fixed F.T.) Zero-shot performance of \textsc{multi} models (\textsc{z.s. multi}) and \textsc{mono} models not trained on the target language. Grey cells indicate full-shot scenarios. All trained on 8,750 examples.}
\label{tab:zero-shot_LLaMA}
\end{table}

\newpage
\section{On the Lack of Extensive Hyperparameter Tuning}
\label{app:hyperparameter_justification}

To keep cross-model and cross-language comparisons fair and interpretable, we deliberately avoided extensive per-model or per-language hyperparameter tuning. Instead, we adopted a single configuration taken from independent prior work and applied it uniformly across all models and settings, with an identical early-stopping rule. This choice limits researcher degrees of freedom and reduces the risk of inadvertently “tuning into” our research hypothesis, which can happen when many parameters are adjusted differently across conditions. While more aggressive tuning could raise the peak performance of individual systems, it would blur causal attribution and undermine the comparability that our study seeks to emphasize. 

\newpage
\onecolumn

\section{Error Analysis} \label{sec:error_analysis}

We perform error analysis to further analyze how, and if multilingual models differ from their monolingual counterparts.

Let:
\label{eq:alignment}
\begin{itemize}
    \item $M$ be the set of \textbf{monolingual models} in the same fine-tuning condition,
    \item $m$ be the model of interest, where $m \in M \cup \{\textsc{multi}\}$,
    \item $\text{Err}(m)$ be the set of incorrect predictions made by model $m$,
    \item $\text{Align}(m, m') = \frac{|\text{Err}(m) \cap \text{Err}(m')|}{|\text{Err}(m)|}$, the proportion of $m$'s errors also made by model $m'$.
\end{itemize}

Then, the \textbf{average proportion of alignment} of model $m$ with the monolingual models is given by:

\begin{equation*}
\text{AvgAlign}(m) =
\begin{cases}
\frac{1}{|M| - 1} \sum\limits_{\substack{m' \in M \\ m' \ne m}} \frac{|\text{Err}(m) \cap \text{Err}(m')|}{|\text{Err}(m)|}, & \text{if } m \in M \\[12pt]
\frac{1}{|M|} \sum\limits_{m' \in M} \frac{|\text{Err}(m) \cap \text{Err}(m')|}{|\text{Err}(m)|}, & \text{if } m = \textsc{multi}
\end{cases}
\end{equation*}

\begin{figure*}
    \centering
    \includegraphics[width=\linewidth]{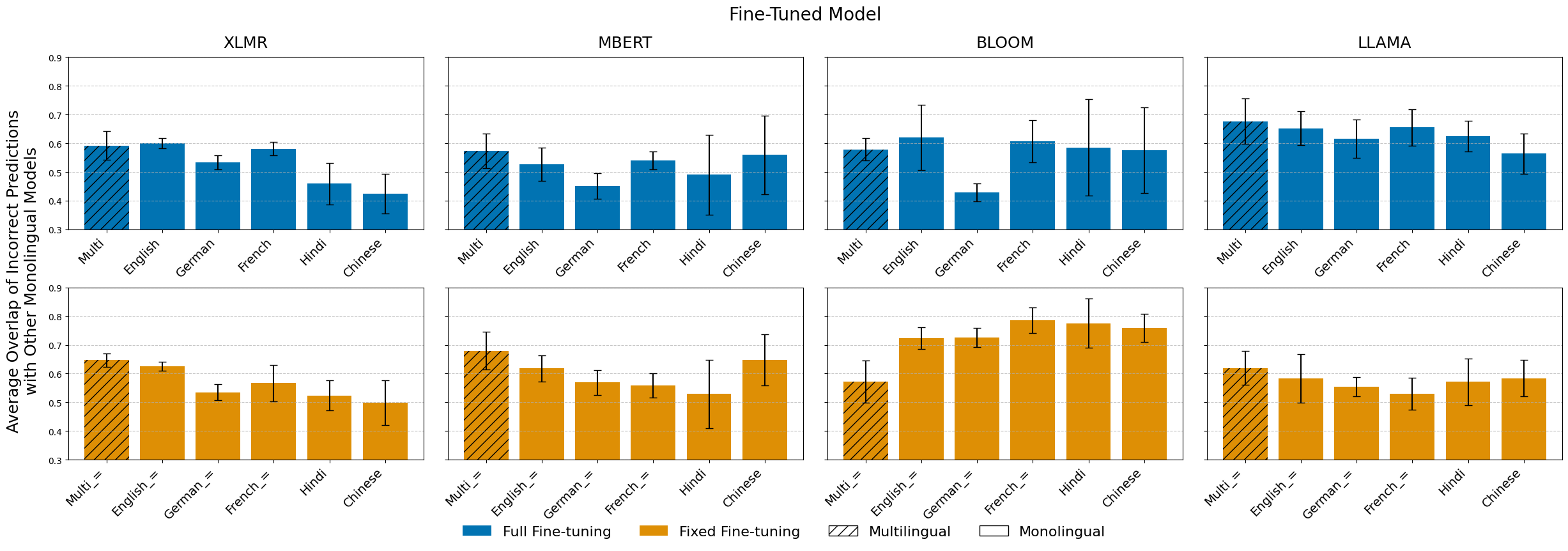}
    \caption{Average proportion of each model's incorrect predictions on the WiC task (across all test files) that are also made by monolingual models in the same fine-tuning condition, as defined in Equation~\ref{eq:alignment}. Multilingual models are indicated with diagonal hatching. The same Hindi and Chinese models are included in both conditions due to their limited fine-tuning size.
}
    \label{fig:incorrect_mono_alignment}
\end{figure*}

\clearpage

\section{Pretraining Sizes of Languages Used in Our Analysis}

\begin{table*}[h!]
    \centering

    \begin{tabular}{llrrll}
    \toprule
    ISO Code & Language & XLM-R & mBERT & BLOOM & MuRIL \\
    \midrule
    AR & Arabic & 3.37 & 0.72 & 4.26 & - \\
    BG & Bulgarian & 4.07 & 0.24 & - & - \\
    BN & Bengali & 2.24 & 0.12 & 2.91 & - \\
    DA & Danish & 3.84 & 0.24 & - & - \\
    DE & German & 4.21 & 1.66 & - & - \\
    EN & English & 5.71 & 2.89 & 6.12 & 3.30 \\
    ES & Spanish & 3.99 & 1.66 & 5.10 & - \\
    ET & Estonian & 1.96 & 0.24 & - & - \\
    EU & Basque & 1.10 & 0.12 & 1.16 & - \\
    FA & Persian & 4.72 & 0.43 & - & - \\
    FI & Finnish & 4.01 & 0.43 & - & - \\
    FR & French & 4.06 & 1.66 & 5.27 & - \\
    HI & Hindi & 3.05 & 0.12 & 3.18 & 1.95 \\
    HR & Croatian & 3.07 & 0.24 & - & - \\
    ID & Indonesian & 5.01 & 0.43 & 2.98 & - \\
    IT & Italian & 3.44 & 1.14 & - & - \\
    JA & Japanese & 4.25 & 1.14 & - & - \\
    KA & Georgian & 2.31 & 0.12 & - & - \\
    KK & Kazakh & 2.00 & 0.12 & - & - \\
    KO & Korean & 4.01 & 0.43 & - & - \\
    LA & Latin & 1.25 & 0.06 & - & - \\
    NL & Dutch & 3.41 & 0.72 & - & - \\
    NO & Norwegian & 3.91 & 0.43 & - & - \\
    RU & Russian & 5.63 & 1.66 & - & - \\
    SV & Swedish & 2.57 & 0.72 & - & - \\
    TR & Turkish & 3.09 & 0.43 & - & - \\
    UR & Urdu & 1.90 & 0.12 & 1.28 & 1.00 \\
    ZH & Simplified Chinese & 3.87 & 1.14 & 5.50 & - \\
    ZH & Traditional Chinese & 2.87 & 1.14 & 0.54 & - \\
    \bottomrule
    \end{tabular}
    \caption{Models' log-transformed pretraining sizes (originally in GB) of languages used in our analysis. \newline Language proportions for LLaMA-3–8B-Instruct's pretraining corpus are not publicly available.}
    \label{tab:pretraining_sizes}
    \footnotetext{The pretraining corpus sizes for languages used to train mBERT are also not publicly reported. For these, we rely on averaged estimates from \citet{wu2020all}, who measured the Wikipedia size for each language included in mBERT's training data.}

\end{table*}

\clearpage
\section{All Results on Monolingual and Cross-Lingual WiC tasks}
\label{app:all_results}

\label{sec:all-results}
\begin{table*}[h]
  \centering
  \small
  \begin{adjustbox}{max width=\textwidth}
\begin{tabular}{|*{21}{c|}} \cline{1-21}
  \multirow{3}{*}{Type} & \multicolumn{1}{|c|}{Dataset $\rightarrow$} & \multicolumn{5}{c|}{MCL} & \multicolumn{13}{c|}{XL} & \multicolumn{1}{c|}{Hindi} \\
    \cline{2-21}
     & \diagbox{Model}{Language} & AR & EN & FR & RU & ZH & BG & DA & DE & EN & ET & FA & FR & HR & IT & JA & KO & NL & ZH & HI  \\ \hline
        \textsc{multi} &\textsc{multi} & \cellcolor{lightyellow}  82.6 & \cellcolor{lightyellow} 88.0 & \cellcolor{lightyellow} 82.6 & \cellcolor{lightyellow} 84.3 & \cellcolor{lightyellow} 85.5 & \cellcolor{lightyellow} 83.0 & \cellcolor{lightyellow} 76.0 & \cellcolor{lightyellow} 83.7 & \cellcolor{lightyellow} 70.4 & \cellcolor{lightyellow} 62.3 & \cellcolor{lightyellow} 66.6 & \cellcolor{lightyellow} 78.0 & \cellcolor{lightyellow} 70.6 & \cellcolor{lightyellow} 72.5 & \cellcolor{lightyellow} 66.7 & \cellcolor{lightyellow} 78.4 & \cellcolor{lightyellow} 77.1 & \cellcolor{lightyellow} 79.4 &  79.3 \\
\hline

    \end{tabular}
    \end{adjustbox}

    \caption{XLM-R models' accuracies on all monolingual WiC tasks. Full-shot conditions where the fine-tuning included the target language are highlighted in yellow.
    }
    \label{tab:xlmr-monolingual}
\end{table*}

\begin{table*}[h]
    \centering
    \tiny
    \begin{adjustbox}{max width=\textwidth}
    \begin{tabular}{|*{16}{c|}}
    \hline
     \multirow{3}{*}{Type} & \multicolumn{1}{|c|}{Dataset $\rightarrow$} & \multicolumn{14}{|c|}{AM$^{2}$iCO}  \\ \cline{2-16}
     & \diagbox{Model}{Language} & AR & BN & DE & EU & FI & ID & JA & KA & KK & KO & RU & TR & UR & ZH  \\ \hline
    \textsc{multi} & \textsc{multi} &  68.0 & 65.7 & 69.3 & 65.4 & 70.0 & 67.8 & 65.8 & 67.2 & 63.2 & 66.9 & 69.9 & 67.4 & 64.5 & 66.6 \\ \hline

        \multirow{5}{*}{\textsc{mono}} & \textsc{german} &  64.5 & 63.9 & 67.8 & 66.6 & 68.3 & 67.1 & 64.4 & 64.2 & 62.5 & 66.9 & 67.7 & 66.4 & 61.2 & 65.9 \\
        & \textsc{french} &  67.7 & 68.0 & 70.0 & 66.7 & 72.5 & 68.2 & 68.3 & 68.9 & 61.8 & 67.9 & 72.4 & 68.2 & 61.5 & 66.5 \\
        & \textsc{english} &  67.4 & 66.1 & 69.9 & 66.2 & 73.0 & 68.7 & 68.5 & 66.5 & 61.8 & 68.9 & 71.8 & 68.7 & 66.0 & 67.6 \\
        & \textsc{hindi} &  58.4 & 51.3 & 60.6 & 55.0 & 60.1 & 54.8 & 61.5 & 52.3 & 52.2 & 61.5 & 60.3 & 58.0 & 55.8 & 59.7 \\
        & \textsc{chinese} &  56.0 & 50.1 & 59.7 & 57.0 & 58.6 & 56.4 & 60.0 & 53.0 & 55.0 & 58.9 & 57.1 & 57.7 & 55.5 & 59.9 \\
\hline
    \end{tabular}
    \end{adjustbox}
    \caption{XLM-R models' accuracies on inter-lingual WiC tasks. Each language in the AM$^2$iCO dataset is paired with English, making the tasks inter-lingual. }
    \label{tab:crosslingual}
\end{table*}

\begin{table*}[h]
    \centering
    \small
    
    \begin{adjustbox}{max width=\textwidth}
    \begin{tabular}{|*{21}{c|}} \cline{1-21}
\multirow{3}{*}{Type} & \multicolumn{1}{c|}{Dataset $\rightarrow$} & \multicolumn{5}{c|}{MCL} & \multicolumn{13}{c|}{XL} & \multicolumn{1}{c|}{Hindi} \\ 

    \cline{2-21}
     & \diagbox{Model}{Language} & AR & EN & FR & RU & ZH & BG & DA & DE & EN & ET & FA & FR & HR & IT & JA & KO & NL & ZH & HI \\ \hline
        \textsc{multi} & \textsc{multi} & \cellcolor{lightyellow}75.1 & \cellcolor{lightyellow}81.4 & \cellcolor{lightyellow}76.2 & \cellcolor{lightyellow}76.7 & \cellcolor{lightyellow}74.8 & \cellcolor{lightyellow}79.5 & \cellcolor{lightyellow}65.9 & \cellcolor{lightyellow}84.8 & \cellcolor{lightyellow}64.1 & \cellcolor{lightyellow}56.2 & \cellcolor{lightyellow}64.4 & \cellcolor{lightyellow}74.6 & \cellcolor{lightyellow}68.6 & \cellcolor{lightyellow}72.0 & \cellcolor{lightyellow}61.2 & \cellcolor{lightyellow}67.1 & \cellcolor{lightyellow}69.7 & \cellcolor{lightyellow}72.2 & 70.7 \\
\hline
    \end{tabular}
    \end{adjustbox}
    \caption{mBERT models' accuracies on all monolingual WiC tasks. Full-shot conditions where the fine-tuning included the target language are highlighted in yellow.}
\end{table*}

\begin{table*}[h]
    \centering
    \small
    \begin{adjustbox}{max width=\textwidth}
    \begin{tabular}{|*{16}{c|}} 
    \hline
     \multirow{3}{*}{Type} & \multicolumn{1}{|c|}{Dataset $\rightarrow$} & \multicolumn{14}{|c|}{AM$^{2}$iCO}  \\ \cline{2-16}
     & \diagbox{Model}{Language} & AR & BN & DE & EU & FI & ID & JA & KA & KK & KO & RU & TR & UR & ZH  \\ \hline
        \textsc{multi} & \textsc{multi} &  60.5 & 61.6 & 64.9 & 59.1 & 64.4 & 62.7 & 60.4 & 59.9 & 57.2 & 59.7 & 64.8 & 61.1 & 57.8 & 60.9 \\ \hline

        \multirow{5}{*}{\textsc{mono}} & \textsc{german} & 56.7 & 60.1 & 61.9 & 59.6 & 61.9 & 61.7 & 59.7 & 58.1 & 59.2 & 57.8 & 62.3 & 61.7 & 57.2 & 58.7 \\ 
        & \textsc{french} & 58.5 & 57 & 62.4 & 61 & 64.4 & 64.1 & 62.6 & 58 & 55 & 59.6 & 64.2 & 62.4 & 57.5 & 60.7 \\ 
        & \textsc{english} & 59.8 & 60 & 63.8 & 60.6 & 63.9 & 64.7 & 61.5 & 61.2 & 59 & 62.4 & 63.9 & 61.6 & 61.5 & 61.5 \\ 
        & \textsc{hindi} & 51.2 & 50.9 & 57.2 & 50.1 & 50.8 & 52.2 & 49.9 & 50.2 & 49.2 & 50.8 & 53.8 & 50.8 & 51 & 50 \\ 
        & \textsc{chinese} & 53 & 52.7 & 59.7 & 50.2 & 52.7 & 55.9 & 52.9 & 51 & 48.8 & 52.9 & 58.8 & 52.8 & 50.7 & 51.5 \\
\hline

    \end{tabular}
    \end{adjustbox}
    \caption{mBERT models' accuracies on inter-lingual WiC tasks. }
\end{table*}

\begin{table*}[h]
    \centering
    \small
    \begin{adjustbox}{max width=\textwidth}
    \begin{tabular}{|*{21}{c|}} \cline{1-21}
    \multirow{3}{*}{Type} & \multicolumn{1}{c|}{Dataset $\rightarrow$} & \multicolumn{5}{c|}{MCL} & \multicolumn{13}{c|}{XL} & \multicolumn{1}{c|}{Hindi} \\ 

    \cline{2-21}
     & \diagbox{Model}{Language} & AR & EN & FR & RU & ZH & BG & DA & DE & EN & ET & FA & FR & HR & IT & JA & KO & NL & ZH & HI  \\ \hline
        \textsc{multi} & \textsc{multi}  & \cellcolor{lightyellow}71.4 & \cellcolor{lightyellow}73.9 & \cellcolor{lightyellow}65.8 & \cellcolor{lightyellow}54.8 & \cellcolor{lightyellow}75.6 & \cellcolor{lightyellow}74.0 & \cellcolor{lightyellow}56.1 & \cellcolor{lightyellow}80.8 & \cellcolor{lightyellow}62.3 & \cellcolor{lightyellow}58.7  & \cellcolor{lightyellow}55.1 & \cellcolor{lightyellow}72.3 & \cellcolor{lightyellow}54.2 & \cellcolor{lightyellow}66.7 & \cellcolor{lightyellow}57.5  & \cellcolor{lightyellow}69.3 & \cellcolor{lightyellow}59.5 & \cellcolor{lightyellow}71.0 & 70.9  \\
\hline
    \end{tabular}
    \end{adjustbox}
    \caption{BLOOM models' accuracies on all monolingual WiC tasks. Full-shot conditions where the fine-tuning included the target language are highlighted in yellow.}
\end{table*}

\begin{table*}[h]
    \centering
    \small
    \begin{adjustbox}{max width=\textwidth}
    \begin{tabular}{|*{16}{c|}} 
    \hline
     \multirow{3}{*}{Type} & \multicolumn{1}{|c|}{Dataset $\rightarrow$} & \multicolumn{14}{|c|}{AM$^{2}$iCO}  \\ \cline{2-16}
     & \diagbox{Model}{Language} & AR & BN & DE & EU & FI & ID & JA & KA & KK & KO & RU & TR & UR & ZH  \\ \hline
        \textsc{multi} & \textsc{multi} &  58.0 & 56.6 & 55.2 & 54.0 & 52.3 & 57.4 & 56.2 & 52.4 & 53.2 & 53.6 & 51.8 & 52.8 & 55.0 & 59.6 \\ \hline

        \multirow{5}{*}{\textsc{mono}} & \textsc{german} & 55.2 & 55.6 & 52 & 54 & 52.6 & 53.4 & 55.5 & 51.8 & 57.2 & 49.3 & 53 & 53.3 & 54 & 55 \\ 
        & \textsc{french} & 57.8 & 54.1 & 55.5 & 51.3 & 50.9 & 58.4 & 53.6 & 49.1 & 50.7 & 51.4 & 51.2 & 50.4 & 57.5 & 60.8 \\ 
        & \textsc{english} & 59.2 & 54.7 & 55.8 & 52.8 & 53.4 & 56.3 & 55.4 & 51.4 & 51.8 & 49.9 & 52.6 & 50.3 & 55.8 & 55.1 \\ 
        & \textsc{hindi} & 50 & 50 & 50 & 49 & 50 & 50.1 & 50.1 & 50.1 & 49.8 & 50.4 & 50.1 & 50 & 50 & 50.4 \\ 
        & \textsc{chinese} & 50 & 50 & 50 & 47.5 & 49.9 & 50 & 51.4 & 50.1 & 49.5 & 49.8 & 50.9 & 49.9 & 50 & 48.2 \\
\hline

    \end{tabular}
    \end{adjustbox}
    \caption{BLOOM models' accuracies on inter-lingual WiC tasks. }
    \label{tab:bloom-crosslingual}
\end{table*}

\begin{table*}[h]
    \centering
    \small
    \begin{adjustbox}{max width=\textwidth}
    \begin{tabular}{|*{21}{c|}} \cline{1-21}
    \multirow{3}{*}{Type} & \multicolumn{1}{c|}{Dataset $\rightarrow$} & \multicolumn{5}{c|}{MCL} & \multicolumn{13}{c|}{XL} & \multicolumn{1}{c|}{Hindi} \\ 

    \cline{2-21}
     & \diagbox{Model}{Language} & AR & EN & FR & RU & ZH & BG & DA & DE & EN & ET & FA & FR & HR & IT & JA & KO & NL & ZH & HI  \\ \hline
     
        \textsc{multi} & \textsc{multi}  &  \cellcolor{lightyellow}71.8 & \cellcolor{lightyellow}85.6 & \cellcolor{lightyellow}76.5 & \cellcolor{lightyellow}78.1 & \cellcolor{lightyellow}80.2 & \cellcolor{lightyellow}69.5 & \cellcolor{lightyellow}67.9 & \cellcolor{lightyellow}77.9 & \cellcolor{lightyellow}71.6 & \cellcolor{lightyellow}62.1 & \cellcolor{lightyellow}79.0 & \cellcolor{lightyellow}74.2 & \cellcolor{lightyellow}69.4 & \cellcolor{lightyellow}70.9 & \cellcolor{lightyellow}66.9 & \cellcolor{lightyellow}73.7 & \cellcolor{lightyellow}75.1 & \cellcolor{lightyellow}72.8 & 58.7
 \\ \hline
        \multirow{5}{*}{\textsc{mono}} & \textsc{german} &  71.5 & 78.9 & 77.0 & 75.9 & 77.9 & 64.3 & 68.9 & \cellcolor{lightyellow} 80.0 & 70.2 & 53.3 & 69.0 & 69.0 & 67.6 & 64.4 & 63.6 & 65.6 & 71.2 & 72.4 & 65.5 \\
        & \textsc{french} &  70.4 & 78.7 & \cellcolor{lightyellow} 72.7 & 71.7 & 75.3 & 62.8 & 67.0 & 69.9 & 69.6 & 53.3 & 69.2 & \cellcolor{lightyellow} 76.5 & 63.0 & 67.6 & 63.1 & 68.6 & 72.7 & 70.3 & 59.7 \\
        & \textsc{english} &  70.3 & \cellcolor{lightyellow} 85.0 & 76.9 & 74.8 & 70.9 & 54.9 & 62.3 & 60.3 & \cellcolor{lightyellow} 67.7 & 50.5 & 63.2 & 63.2 & 55.1 & 57.1 & 53.2 & 55.6 & 59.8 & 65.4 & 62.6 \\ 
        & \textsc{hindi} &  64.8 & 72.7 & 66.4 & 68.7 & 69.4 & 56.4 & 58.3 & 59.8 & 64.1 & 50.8 & 67.6 & 56.8 & 60.0 & 58.4 & 57.5 & 58.2 & 62.8 & 64.7 & \cellcolor{lightyellow} 72.9 \\
\hline
    \end{tabular}
    \end{adjustbox}
    \caption{LLaMA models' accuracies on all monolingual WiC tasks. Full-shot conditions where the fine-tuning included the target language are highlighted in yellow.}
\end{table*}

\begin{table*}[h]
    \centering
    \small
    \begin{adjustbox}{max width=\textwidth}
    \begin{tabular}{|*{16}{c|}} 
    \hline
     \multirow{3}{*}{Type} & \multicolumn{1}{|c|}{Dataset $\rightarrow$} & \multicolumn{14}{|c|}{AM$^{2}$iCO}  \\ \cline{2-16}
     & \diagbox{Model}{Language} & AR & BN & DE & EU & FI & ID & JA & KA & KK & KO & RU & TR & UR & ZH  \\ \hline

        \textsc{multi} & \textsc{multi} & 61.1 & 59.7 & 63.5 & 58.1 & 60.0 & 59.8 & 59.7 & 57.4 & 56.0 & 61.7 & 60.3 & 58.6 & 55.2 & 61.5 \\ \hline

        \multirow{5}{*}{\textsc{mono}} & \textsc{german} &  59.9 & 55.0 & 60.6 & 55.2 & 57.7 & 57.3 & 60.0 & 55.0 & 56.8 & 60.2 & 58.3 & 58.3 & 54.5 & 57.3 \\ 
        & \textsc{french} &  58.5 & 58.0 & 62.5 & 57.1 & 63.6 & 58.8 & 59.4 & 56.5 & 53.2 & 60.5 & 60.3 & 60.2 & 57.2 & 59.0 \\ 
        & \textsc{english} &  61.9 & 61.0 & 63.9 & 57.6 & 63.9 & 60.1 & 62.0 & 59.5 & 57.0 & 63.6 & 62.3 & 60.1 & 59.5 & 62.1 \\
        & \textsc{hindi} &  59.8 & 59.7 & 62.1 & 56.1 & 60.0 & 59.8 & 59.1 & 55.8 & 55.2 & 59.5 & 60.5 & 57.0 & 55.5 & 58.6 \\
        & \textsc{chinese} &  59.0 & 56.4 & 62.2 & 55.4 & 59.4 & 60.1 & 58.8 & 53.2 & 53.5 & 56.3 & 59.8 & 59.3 & 55.5 & 58.5 \\
\hline

    \end{tabular}
    \end{adjustbox}
    \caption{LLaMA models' accuracies on inter-lingual WiC tasks. }
    \label{tab:bloom-crosslingual}
\end{table*}

\begin{table*}[h]
    \centering
    \small
    \begin{adjustbox}{max width=\textwidth}
    \begin{tabular}{|*{21}{c|}} \cline{1-21}
    \multirow{3}{*}{Type} & \multicolumn{1}{c|}{Dataset $\rightarrow$} & \multicolumn{5}{c|}{MCL} & \multicolumn{13}{c|}{XL} & \multicolumn{1}{c|}{Hindi} \\ 

    \cline{2-21}
     & \diagbox{Model}{Language} & AR & EN & FR & RU & ZH & BG & DA & DE & EN & ET & FA & FR & HR & IT & JA & KO & NL & ZH & HI  \\ \hline
    \multirow{5}{*}{\textsc{mono}} & \textsc{hindi}  &  56.8 & 56.6 & 50.9 & 50.1 & 49.2 & 50.3 & 52.7 & 49.5 & 56.3 & 51.5 & 50.5 & 51.3 & 50.5 & 57.4 & 52.7 & 53.0 & 53.5 & 48.8 & \cellcolor{lightyellow} 86.5 \\ 
    & \textsc{english} &  56.2 & \cellcolor{lightyellow} 86.6 & 64.5 & 50.7 & 49.1 & 50.2 & 52.4 & 52.5 & \cellcolor{lightyellow} 70.4 & 51.0 & 49.1 & 56.0 & 46.3 & 53.4 & 50.0 & 50.0 & 51.0 & 50.1 & 81.4 \\ 
    & \textsc{french} &  52.7 & 61.2 & \cellcolor{lightyellow} 56.7 & 50.3 & 50.0 & 52.8 & 56.1 & 48.8 & 58.7 & 54.9 & 51.5 & \cellcolor{lightyellow} 50.9 & 50.7 & 54.2 & 50.8 & 59.6 & 53.8 & 54.0 & 64.0 \\ 
    & \textsc{german} &  55.3 & 68.5 & 62.7 & 51.0 & 51.2 & 49.9 & 56.0 & \cellcolor{lightyellow} 80.0 & 62.0 & 55.1 & 56.5 & 57.4 & 48.8 & 55.2 & 51.1 & 50.0 & 56.4 & 51.0 & 69.8 \\ 
    & \textsc{chinese} &  50.0 & 50.2 & 52.8 & 50.9 & \cellcolor{lightyellow} 54.3 & 49.9 & 51.4 & 51.6 & 51.7 & 52.1 & 51.9 & 51.6 & 53.2 & 55.4 & 50.6 & 58.4 & 51.5 & \cellcolor{lightyellow} 52.1 & 56.6 \\
\hline
    \end{tabular}
    \end{adjustbox}
    \caption{MuRIL models' accuracies on all monolingual WiC tasks. Full-shot conditions where the fine-tuning included the target language are highlighted in yellow.}
\end{table*}

\begin{table*}[h]
    \centering
    \small
    \begin{adjustbox}{max width=\textwidth}
    \begin{tabular}{|*{16}{c|}} 
    \hline
     \multirow{3}{*}{Type} & \multicolumn{1}{|c|}{Dataset $\rightarrow$} & \multicolumn{14}{|c|}{AM$^{2}$iCO}  \\ \cline{2-16}
     & \diagbox{Model}{Language} & AR & BN & DE & EU & FI & ID & JA & KA & KK & KO & RU & TR & UR & ZH  \\ \hline
    \multirow{5}{*}{\textsc{mono}} & \textsc{hindi} &  49.6 & 50.3 & 50.8 & 49.5 & 52.1 & 48.5 & 47.5 & 49.9 & 49.0 & 50.8 & 50.6 & 50.2 & 50.7 & 50.0 \\
    & \textsc{english} &  53.7 & 71.7 & 65.3 & 55.9 & 57.3 & 59.8 & 49.8 & 49.0 & 49.0 & 49.3 & 51.7 & 54.8 & 70.2 & 48.7 \\
    & \textsc{french} &  49.5 & 51.1 & 51.6 & 48.8 & 49.7 & 49.9 & 50.0 & 48.1 & 50.7 & 47.5 & 51.9 & 51.4 & 51.5 & 50.2 \\
    & \textsc{german} &  52.6 & 57.6 & 56.9 & 54.8 & 54.4 & 56.2 & 51.1 & 52.2 & 51.8 & 50.2 & 53.7 & 52.5 & 53.8 & 51.1 \\ 
    & \textsc{chinese} &  50.0 & 50.1 & 50.0 & 50.0 & 49.9 & 50.0 & 50.0 & 50.0 & 50.0 & 50.0 & 49.9 & 50.0 & 50.0 & 50.0 \\ \hline
    \textsc{multi} & \textsc{multi} &  49.6 & 47.7 & 50.4 & 48.5 & 49.5 & 49.3 & 47.7 & 48.9 & 51.5 & 50.2 & 49.1 & 49.3 & 49.2 & 48.1  \\ \hline     
    \end{tabular}
    \end{adjustbox}
    \caption{MuRIL models' accuracies on inter-lingual WiC tasks. }
    \label{tab:bloom-crosslingual}
\end{table*}

\clearpage

\clearpage
\section{Performance of Pretrained Models without Fine-Tuning}
\label{app:paramfree}

It is logical to ask whether pretrained multilingual language models already separate word senses in a no-fine-tuning setting. Because contextual embeddings integrate sentence context via attention, one might expect embeddings of \emph{mole} in “burrowing mammal” vs. “skin blemish” to diverge more than two occurrences of the same sense, even without fine-tuning. To probe this, we evaluate pretrained models without task-specific training by classifying sentence pairs using cosine distance between target-word embeddings, consistent with our main protocol. 

\paragraph{Threshold selection and considerations.}
In our main experiment setup, each fine-tuned encoder uses a cosine-distance decision threshold set on dev data from its fine-tuning language (or multilingual dev for MULTI models). For encoders without fine-tuning, no model-specific dev data exists, leaving only flawed alternatives: (A) calibrate once on pooled multilingual dev data (introducing mixture bias), (B) tune separately on each test set (test leakage, inflated scores), or (C) calibrate per language (uses target-language supervision unavailable to the pretrained encoders). We adopt Option~A as the least biased feasible choice. We sweep candidate thresholds on the pooled dev data to maximize accuracy and then fix that single global threshold for each encoder across all datasets. This avoids test leakage and keeps calibration constant, but the global threshold inevitably reflects mixture statistics and tailors to the average of all the data, and can favor high-resource languages. Accordingly, these numbers are a \emph{sanity check} for learning rather than directly comparable performance estimates.

\begin{figure*}[t]
    \centering
    \includegraphics[width=1\linewidth]{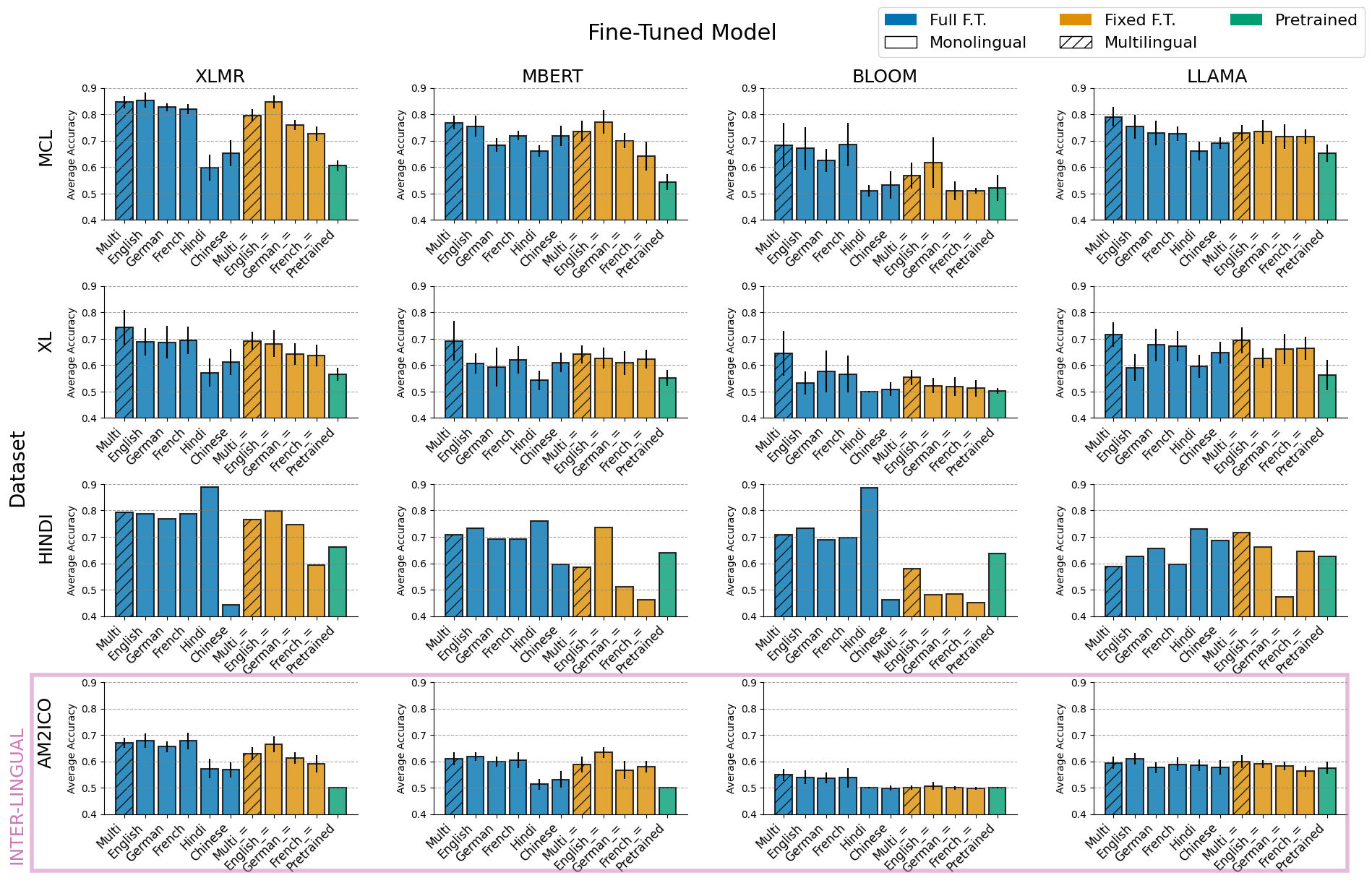}
    \caption{
    Mean accuracy with standard deviation for multilingual and monolingual models on WiC datasets, showing both pretrained (no-fine-tuning) and fine-tuned performance. Color encodes whether fine-tuning used the full training set, a subsampled portion, or whether the pretrained models were used off-the-shelf. Hindi and Chinese appear only in the full-data condition due to limited size. The Hindi dataset has no error bars because it is a single-language dataset.}
    \label{fig:mlcl_vs_mono_ft_w_pretraining}
\end{figure*}

\paragraph{Results.} Figure~\ref{fig:mlcl_vs_mono_ft_w_pretraining} shows that XLM-R, mBERT, and BLOOM consistently benefit from fine-tuning across nearly all conditions, with the few exceptions concentrated in low-data fine-tuning settings (e.g., Chinese or Fixed F.T.) when evaluated on Hindi. These gains indicate that fine-tuning substantially improves sense discrimination beyond what is present in the pretrained geometry.

For LLaMA, the dynamics differ slightly. The generative model is prompted (see Figure~\ref{fig:llama_prompt} for the prompt) rather than based on a cosine-distance threshold. Its zero-shot instruction-following yields a stronger baseline than the pretrained encoders on most tasks; accordingly, despite using the same WiC training data, its absolute gains from fine-tuning are smaller. This reflects an architectural/interface contrast—encoders benefit from shaping the embedding space and a calibrated decision threshold, whereas LLaMA already executes the task from the prompt and changes less with additional supervision.

We include full results for the no-fine-tuning baselines in Table~\ref{tab:pretrained-multilingual}. These baselines show that sense discrimination improves with fine-tuning; however, they are not intended for head-to-head comparison with fine-tuned models due to the necessarily different threshold calibration.

\begin{table*}[h]
    \centering
    \small
    \begin{adjustbox}{max width=\textwidth}
    \begin{tabular}{|*{20}{c|}} \cline{1-20}
    Dataset $\rightarrow$ & \multicolumn{5}{c|}{MCL} & \multicolumn{13}{c|}{XL} & \multicolumn{1}{c|}{Hindi} \\ 
    \hline
     \diagbox{Model}{Language} & AR & EN & FR & RU & ZH & BG & DA & DE & EN & ET & FA & FR & HR & IT & JA & KO & NL & ZH & HI  \\ \hline
     
     \textsc{XLM-R}  & 62.2 & 62.2 & 61.6 & 58.7 & 58.0 & 58.5 & 60.6 & 56.1 & 56.1 & 53.1 & 53.9 & 53.1 & 57.0 & 54.8 & 56.9 & 55.7 & 58.7 & 60.0 & 66.1 \\ 
    \textsc{mBERT} & 56.2 & 58.5 & 51.2 & 52.1 & 53.7 & 58.1 & 54.2 & 54.4 & 52.8 & 54.7 & 55.2 & 52.6 & 54.7 & 50.3 & 56.4 & 56.8 & 62.6 & 53.1  & 63.8\\ 
    \textsc{BLOOM} & 61.1 & 50.0 & 50.0 & 50.0 & 50.0 & 49.9 & 50.0 & 49.0 & 50.0 & 50.1 & 49.9 & 50.0 & 50.1 & 49.7 & 50.3 & 50.0 & 50.0 & 53.9 & 63.6\\ 
    \textsc{LLaMA} & 67.2 & 64.2 & 65.4 & 58.0 & 62.8 & 55.4 & 63.2 & 56.9 & 55.7 & 58.4 & 57.9 & 53.8 & 53.4 & 52.8 & 46.5 & 54.1 & 53.7 & 70.6 & 62.5\\ 
    \textsc{MuRIL} & 50.0 & 49.9 & 57.4 & 51.8 & 50.4 & 49.8 & 49.9 & 48.5 & 50.8 & 50.6 & 56.1 & 58.2 & 49.7 & 50.0 & 56.2 & 50.2 & 53.2 & 50.4 & 59.8\\ \hline

    \end{tabular}
    \end{adjustbox}
    \caption{Pretrained models' accuracies on all monolingual WiC tasks. }
    \label{tab:pretrained-multilingual}
\end{table*}

\begin{table*}[h]
    \centering
    \small
    \begin{adjustbox}{max width=\textwidth}
    \begin{tabular}{|*{15}{c|}} 
    \hline
    \multicolumn{1}{|c|}{Dataset $\rightarrow$} & \multicolumn{14}{|c|}{AM$^{2}$iCO}  \\ \hline
    \diagbox{Model}{Language} & AR & BN & DE & EU & FI & ID & JA & KA & KK & KO & RU & TR & UR & ZH  \\ \hline
    \textsc{XLM-R} & 50.0 & 50.0 & 50.0 & 50.0 & 50.0 & 50.0 & 50.0 & 50.0 & 50.0 & 50.0 & 49.9 & 50.1 & 50.0 & 50.0 \\
    \textsc{mBERT} & 50.0 & 50.0 & 50.1 & 50.1 & 50.0 & 50.0 & 50.0 & 50.0 & 50.0 & 50.0 & 49.9 & 49.9 & 50.0 & 50.0 \\
    \textsc{BLOOM} & 50.0 & 50.0 & 50.0 & 50.8 & 49.9 & 50.1 & 50.2 & 50.1 & 49.8 & 49.9 & 49.7 & 49.9 & 50.0 & 49.8 \\
    \textsc{LLaMA} & 58.3 & 55.7 & 61.6 & 55.3 & 57.3 & 59.3 & 58.7 & 54.7 & 53.2 & 57.9 & 57.7 & 56.8 & 59.0 & 60.3 \\
    \textsc{MuRIL} & 50.5 & 49.7 & 50.3 & 50.0 & 49.8 & 50.6 & 49.7 & 50.6 & 50.0 & 49.6 & 49.8 & 50.1 & 52.2 & 50.0 \\ \hline

    \end{tabular}
    \end{adjustbox}
    \caption{Pretrained models' accuracies on inter-lingual WiC tasks. }
    \label{tab:pretrained-crosslingual}
\end{table*}

\clearpage
\onecolumn
\section{Performance of XLM-R, mBERT and BLOOM-Based Models on LSCD Tasks}

MuRIL models were excluded from this evaluation due to poor cross-lingual performance, resulting from a lack of pretraining on most of the target languages. LLaMA models were also not evaluated, as the LSCD task format is not well-suited to generative architectures.

\begin{table*}[h]
\centering
\small
\begin{tabular}{|c|c|c|c|c|c|c|c|c|}
\hline
Language & Metric & \textsc{xl-lexeme} & \textsc{multi} & \textsc{german} & \textsc{french} & \textsc{english} & \textsc{hindi} & \textsc{chinese} \\
\hline
\multirow{2}{*}{English}
  & APD & \underline{.757} & .703 & \underline{.737} & .681 & \textbf{.772} & .436 & .673 \\
  & PRT & \underline{.495} & \underline{.492} & .241 & .337 & \textbf{.535} & .363 & .367 \\
\hline
\multirow{2}{*}{German}
  & APD & \textbf{.873} & \underline{.863} & \underline{.841} & \underline{.867} & \underline{.844} & .635 & .641 \\
  & PRT & \underline{.881} & \textbf{.890} & .829 & .831 & \underline{.873} & .755 & .682 \\
\hline
\multirow{2}{*}{Swedish}
  & APD & \textbf{.755} & .801 & \underline{.754} & .618 & \underline{.724} & .480 & .489 \\
  & PRT & \textbf{.678} & \underline{.673} & .522 & .138 & .627 & .277 & .332 \\
\hline
\multirow{2}{*}{Latin}
  & APD & -.035 & .117 & \textbf{.161} & \underline{.136} & \underline{.135} & -.177 & .091 \\
  & PRT & \underline{.467} & .392 & .445 & .405 & .429 & \textbf{.512} & \textbf{.512} \\
\hline
\multirow{2}{*}{Spanish}
  & APD & \underline{.665} & \underline{.696} & \underline{.670} & \underline{.664} & \textbf{.711} & .354 & .383 \\
  & PRT & .633 & \textbf{.698} & \underline{.655} & \underline{.649} & .643 & .355 & .267 \\
\hline
\multirow{2}{*}{Chinese}
  & APD & \underline{.734} & .652 & .649 & .499 & \textbf{.737} & .524 & .593 \\
  & PRT & \underline{.702} & \textbf{.708} & .623 & .578 & \underline{.684 }& .552 & .432 \\
\hline
\multirow{2}{*}{Norwegian$_1$}
  & APD & .668 & \underline{.729} & .638 & .697 & \textbf{.777} & .525 & .400 \\
  & PRT & .769 & .784 & .730 & .740 & \textbf{.845} & .551 & .435 \\
\hline
\multirow{2}{*}{Norwegian$_2$}
  & APD & \underline{.634} & \textbf{.655} & .604 & .580 & \underline{.645} & .433 & .439 \\
  & PRT & .532 & \underline{.583} & .557 & .525 & \textbf{.636} & .337 & .396 \\
\hline
\multirow{2}{*}{Average}
  & APD & \underline{.631} & \underline{.652} & \underline{.632} & .593 & \textbf{.668} & .401 & .464 \\
  & PRT & \underline{.645} & \underline{.652} & .575 & .525 & \textbf{.659} & .463 & .428 \\
\hline
\end{tabular}
\caption{Spearman correlations of XLM-R models' APD and PRT scores with graded semantic change scores across LSCD tasks, transposed to show languages as rows. Best scores are bolded; scores within 0.05 of the best are underlined.}
\label{tab:XLMR-lscd}
\end{table*}

\begin{table*}[h]
\centering
\small
\begin{tabular}{|c|c|c|c|c|c|c|c|}
\hline
Language & Metric & \textsc{multi} & \textsc{german} & \textsc{french} & \textsc{english} & \textsc{hindi} & \textsc{chinese} \\
\hline
\multirow{2}{*}{English}
  & APD & \textbf{.754} & .566 & .551 & \underline{.711} & .506 & .684 \\
  & PRT & \textbf{.470} & \underline{.397} & \underline{.393} & \underline{.434} & .309 & .328 \\
\hline
\multirow{2}{*}{German}
  & APD & \underline{.760} & .719 & \underline{.772} & \textbf{.810} & .551 & \underline{.783} \\
  & PRT & \underline{.834} & .759 & \textbf{.846} & \underline{.845} & .550 & .705 \\
\hline
\multirow{2}{*}{Swedish}
  & APD & .377 & .473 & .002 & .437 & \underline{.523} & \textbf{.541} \\
  & PRT & .082 & .145 & -.504 & .266 & .207 & \textbf{.366} \\
\hline
\multirow{2}{*}{Latin}
  & APD & \textbf{.181} & -.092 & -.058 & -.158 & \underline{.163} & -.190 \\
  & PRT & .298 & .339 & .235 & .038 & \textbf{.462} & \underline{.412} \\
\hline
\multirow{2}{*}{Spanish}
  & APD & \underline{.629} & .500 & .531 & \textbf{.651} & .452 & .536 \\
  & PRT & \textbf{.636} & .462 & .546 & \underline{.617} & .332 & .407 \\
\hline
\multirow{2}{*}{Chinese}
  & APD & \underline{.654} & .554 & .386 & \textbf{.668} & \underline{.651} & .619 \\
  & PRT & \textbf{.690} & .623 & .493 & \underline{.686} & .427 & .529 \\
\hline
\multirow{2}{*}{Norwegian$_1$}
  & APD & .596 & .515 & .561 & .638 & .631 & \textbf{.700} \\
  & PRT & \underline{.724} & \underline{.698} & .580 & \textbf{.748} & .558 & .601 \\
\hline
\multirow{2}{*}{Norwegian$_2$}
  & APD & \textbf{.601} & \underline{.596} & .464 & \textbf{.604} & .412 & .581 \\
  & PRT & .474 & \underline{.527} & .260 & \textbf{.535} & .238 & .421 \\
\hline
\multirow{2}{*}{Average}
  & APD & \textbf{.569} & .479 & .401 & \underline{.545} & .486 & \underline{.532} \\
  & PRT & \textbf{.526} & \underline{.494} & .356 & \underline{.521} & .385 & .471 \\
\hline
\end{tabular}
\caption{Spearman correlations of mBERT models' APD and PRT scores with graded semantic change scores across LSCD tasks, transposed to show languages as rows. Best scores from fine-tuned models are bolded; scores within 0.05 of the best are underlined.}
\label{tab:mbert-lscd}
\end{table*}

\begin{table*}[h]
\centering
\small
\begin{tabular}{|c|c|c|c|c|c|c|c|}
\hline
Language & Metric & \textsc{multi} & \textsc{german} & \textsc{french} & \textsc{english} & \textsc{hindi} & \textsc{chinese} \\
\hline
\multirow{2}{*}{English}
  & APD & \textbf{.689} & .463 & \underline{.657} & \underline{.656} & .176 & .291 \\
  & PRT & \underline{.469} & .293 & \underline{.459} & \textbf{.487} & .215 & .342 \\
\hline
\multirow{2}{*}{German}
  & APD & \textbf{.634} & \underline{.606} & .244 & .410 & .075 & .144 \\
  & PRT & \underline{.720} & \textbf{.723} & .474 & .564 & .274 & .499 \\
\hline
\multirow{2}{*}{Swedish}
  & APD & .278 & \textbf{.399} & .303 & .311 & .276 & .132 \\
  & PRT & .287 & \textbf{.326} & .219 & -.052 & .134 & -.041 \\
\hline
\multirow{2}{*}{Latin}
  & APD & -.068 & -.104 & \textbf{.081} & -.197 & -.230 & -.173 \\
  & PRT & \textbf{.367} & .221 & \underline{.323} & \underline{.314} & .125 & \underline{.310} \\
\hline
\multirow{2}{*}{Spanish}
  & APD & .501 & .370 & .474 & \textbf{.599} & .326 & .230 \\
  & PRT & \underline{.489} & .372 & .450 & \textbf{.521} & .284 & .178 \\
\hline
\multirow{2}{*}{Chinese}
  & APD & \underline{.550} & .467 & \underline{.570} & \textbf{.592} & .216 & .299 \\
  & PRT & \underline{.546} & .381 & .506 & \textbf{.568} & .395 & .159 \\
\hline
\multirow{2}{*}{Norwegian$_1$}
  & APD & .278 & .289 & .131 & \textbf{.345} & -.002 & .051 \\
  & PRT & .251 & \textbf{.512} & -.015 & .354 & .240 & \underline{.363} \\
\hline
\multirow{2}{*}{Norwegian$_2$}
  & APD & \textbf{.395} & .193 & \underline{.237} & .089 & -.109 & -.046 \\
  & PRT & \textbf{.264} & .216 & \underline{.261} & .041 & -.303 & -.118 \\
\hline
\multirow{2}{*}{Average}
  & APD & \textbf{.407} & .335 & .337 & .351 & .091 & .116 \\
  & PRT & \textbf{.424} & \underline{.380} & .335 & .350 & .170 & .211 \\
\hline
\end{tabular}
\caption{Spearman correlations of BLOOM models' APD and PRT scores with graded semantic change scores across LSCD tasks, transposed to show languages as rows. Best scores from fine-tuned models are bolded; scores within 0.05 of the best are underlined.}
\label{tab:bloom-lscd}
\end{table*}

\clearpage
\twocolumn
\section{LLaMA}

\subsection{LLaMA outperformed by XLM-R}

\begin{figure*}[!ht]
    \centering
    \includegraphics[width=0.7\linewidth]{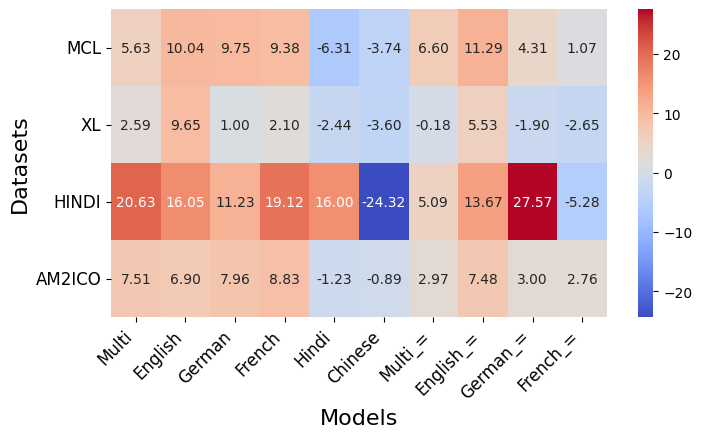}
    \caption{Differences between the per-dataset average accuracies of XLM-R based models and LLaMA-based models trained on the same data in both Full and Fixed Fine-Tuning conditions.}
    \label{fig:xlmr-llama}
\end{figure*}

Figure~\ref{fig:xlmr-llama} shows that XLM-R generally outperforms LLaMA across most datasets under both full and fixed fine-tuning conditions, except when the models are trained on Hindi or Chinese. This can be attributed to LLaMA's instruction-following pretraining, which enables it to better generalize from limited examples, which is particularly valuable in low-resource settings. In contrast, XLM-R lacks such capabilities and must infer both the task and the language from sparse training data. This issue is exacerbated by the fact that LLaMA was provided with prompts, which were consistently in English due to tokenizer limitations, while XLM-R had no explicit indication of either the task or input language during training or evaluation except for the sparse training data.

While we include LLaMA in our evaluation to provide a comparison with recent large language models (LLMs) and instruction-tuned architectures, our results suggest that encoder-based models like XLM-R remain more effective for in-context polysemy disambiguation. Despite the generative capabilities of LLaMA, it underperforms on these tasks in most languages compared to XLM-R, particularly when ample training data is available. This highlights the continued relevance of encoder-based models, which demonstrate stronger task-specific performance and greater cross-lingual transfer.

\section{LLaMA Fine-Tuning Parameters}

\begin{table}[!htb]
\centering
\resizebox{0.5\textwidth}{!}{
\begin{tabular}{|lc|}
\hline
\multicolumn{2}{|c|}{\textbf{Fine-tuning details}}                                                                                                                  \\ \hline
\multicolumn{1}{|c|}{pre-trained LLMs}            & \begin{tabular}[c]{@{}c@{}} \href{https://huggingface.co/meta-llama/Meta-Llama-3-8B-Instruct}{\textit{Meta-Llama-3-8B-Instruct}}\end{tabular}                                                                   \\ \hline
\multicolumn{1}{|c|}{GPUs}                        & NVIDIA A40 (48GB)                                                                                                  \\ \hline
\multicolumn{1}{|c|}{PEFT}                        & LoRA                                                                                                            \\ \hline
\multicolumn{1}{|c|}{LoRA dropout}                & 0.1                                                                                                             \\ \hline
\multicolumn{1}{|c|}{Weight decay}                & 0.001                                                                                                            \\ \hline
\multicolumn{1}{|c|}{Learning rate}               & 1e-4                                                                                                            \\ \hline
\multicolumn{1}{|c|}{LoRArank}                   & 128                                                                                                             \\ \hline
\multicolumn{1}{|c|}{LoRAalpha}                  & 256                                                                                                             \\ \hline
\multicolumn{1}{|c|}{Warmup ratio}                & 0.05                                                                                                            \\ \hline
\multicolumn{1}{|c|}{Num train epochs}            & 3                                                                                                              \\ \hline
\multicolumn{1}{|c|}{Gradient accumulation steps} & 4                                                                                                               \\ \hline
\multicolumn{1}{|c|}{Max seq. length}             & 512                                                                                                             \\ \hline
\multicolumn{1}{|c|}{Batch size}                  & 8                                                                                                              \\ \hline
\multicolumn{1}{|c|}{Optimizer}                   & paged\_adamw\_8bit                                                                                              \\ \hline
\multicolumn{1}{|c|}{LoRA target modules}         & \begin{tabular}[c]{@{}c@{}}q\_proj, v\_proj, k\_proj, o\_proj, \\ gate\_proj, up\_proj, down\_proj\end{tabular} \\ \hline
\end{tabular}}
\caption{Settings and parameters for fine-tuning Llama3Instruct.}
\label{tab:fine-params}
\end{table}

\subsection{Prompts and Response Parsing} \label{sec:llama_details}
\begin{figure*}[t]
\centering
\begin{quote}
\textbf{System message:}

\small
\texttt{Act as an expert lexicographer: determine if a given word has the same sense in two sentences, and respond with 1 if the sense is the same, or 0 if it is different.}

\vspace{1em}
\textbf{User message:}

\small
\begin{verbatim}
Determine if "{target_word}" has the same meaning in these two sentences.
You MUST reply with ONLY:
1 — if the meaning is identical.
0 — if the meaning differs.

Sentence 1: "{sentence1}"
Sentence 2: "{sentence2}"

Answer (ONLY 1 or 0):
\end{verbatim}
\end{quote}
\caption{System and user prompt used to adapt the WiC task for LLaMA.}
\label{fig:llama_prompt}
\end{figure*}

We adapted the WiC task to a generative setting for LLaMA by using the prompt shown in Figure~\ref{fig:llama_prompt}. This prompt format was used consistently during both training and evaluation. For training, the prompt was followed by the correct binary label (``1'' or ``0'') as the target output. During evaluation, the model generated this label based on the prompt alone. We evaluated the pretrained model using several alternative prompt formulations on evaluation in all languages, ultimately selecting the one that yielded the highest overall performance.

To extract model predictions, we used the following simple rule-based parser:

\begin{verbatim}
if "1" in output and "0" not in output:
    return "1"
elif "0" in output and "1" not in output:
    return "0"
else:
    return None # counts as incorrect
\end{verbatim}

In practice, nearly all outputs followed the expected format, with the vast majority consisting of a single ``1'' or ``0''.

\end{document}